%% file: acl_latex.tex
\documentclass[11pt]{article}

\usepackage[preprint]{acl}

\usepackage{times}
\usepackage{latexsym}

\usepackage[T1]{fontenc}

\usepackage[utf8]{inputenc}

\usepackage{microtype}

\usepackage{inconsolata}

\usepackage{graphicx}

\usepackage{multirow}
\usepackage{enumitem}
\usepackage{amsmath}
\usepackage{amssymb}
\usepackage{booktabs}

\usepackage{colortbl} 
\usepackage{hyperref}
\usepackage{subcaption}
\usepackage{ulem}
\usepackage{wrapfig}
\usepackage{fontawesome}
\usepackage{bbm}
\usepackage{pifont}
\newcommand{\tablestyle}[2]{\setlength{\tabcolsep}{#1}\renewcommand{\arraystretch}{#2}\centering\footnotesize}

\definecolor{LightCyan}{RGB}{232,241,255}
\definecolor{my_green}{RGB}{51,102,0}
\definecolor{my_red}{RGB}{204, 0, 0}
\definecolor{COLOR_MEAN}{HTML}{f0f0f0}
\definecolor{category-S1}{HTML}{F5C297}
\definecolor{category-S2}{HTML}{EAC8DA}
\definecolor{category-S3}{HTML}{C4C5E8}
\definecolor{category-S4}{HTML}{B2D5DF}
\definecolor{category-S5}{HTML}{BFE1B6}
\definecolor{lightred}{RGB}{251,49,153}
\renewcommand{\checkmark}{\textcolor{my_green}{\ding{51}}} 
\newcommand{\crossmark}{\textcolor{my_red}{\ding{55}}} 
\usepackage{algorithm}
\usepackage{algpseudocode}
\title{AgentHallu: Benchmarking Automated Hallucination Attribution of LLM-based Agents}

\author{Xuannan Liu$^{1}$ \quad
  Xiao Yang$^{2}$ \quad
  Zekun Li$^{3}$ \quad
  Peipei Li$^{1}$ \quad
  Ran He$^{4}$ \quad \\
  $^{1}$ Beijing University of Posts and Telecommunications \quad \\
$^{2}$ Department of Computer Science \& Technology, Tsinghua University \\
$^{3}$ University of California, Santa Barbara \quad \\
$^{4}$ Center for Research on Intelligent Perception and Computing, NLPR, CASIA
\\
\textcolor{lightred}{ \url{https://liuxuannan.github.io/AgentHallu.github.io/}}
  }

\begin{document}
\maketitle

\input{Sec/0_abstract}

\input{Sec/1_intro}

\input{Sec/2_related_work}

\input{Sec/3_method}

\input{Sec/4_experiment}

\input{Sec/5_conclusion}

\input{Sec/6_limitation}


\bibliography{custom}

\clearpage
\appendix

\begin{center}
{\Large \textbf{Appendix}}
\end{center}

\setcounter{section}{0}
\renewcommand{\thesection}{\Alph{section}}
\tableofcontents

\clearpage

\input{Sec/7_appendix}

\end{document}

%% file: Sec/0_abstract.tex
\begin{abstract}
As LLM-based agents operate over sequential multi-step reasoning, hallucinations arising at intermediate steps risk propagating along the trajectory, thus degrading overall reliability.
Unlike hallucination detection in single-turn responses, diagnosing hallucinations in multi-step workflows requires identifying which step causes the initial divergence. To fill this gap, we propose a new research task, \textbf{automated hallucination attribution} of LLM-based agents, aiming to identify the step responsible for the hallucination and explain why. To support this task, we introduce AgentHallu, a comprehensive benchmark with: (1) 693 high-quality trajectories spanning 7 agent frameworks and 5 domains, (2) a hallucination taxonomy organized into 5 categories (Planning, Retrieval, Reasoning, Human-Interaction, and Tool-Use) and 14 sub-categories, and (3) multi-level annotations curated by humans, covering binary labels, hallucination-responsible steps, and causal explanations.
We evaluate 13 leading models, and results show the task is challenging even for top-tier models (like GPT-5, Gemini-2.5-Pro). 
The best-performing model achieves only 41.1\% step localization accuracy, where tool-use hallucinations are the most challenging at just 11.6\%.
We believe AgentHallu will catalyze future research into developing robust, transparent, and reliable agentic systems.
\end{abstract}

%% file: Sec/1_intro.tex
\section{Introduction}
\label{sec:intro}
Large Language Models (LLMs)~\cite{chatgpt2025, comanici2025gemini} have been increasingly deployed into autonomous agents to tackle complex tasks~\cite{mialon2023gaia,yang2024swe,zheng2024gpt}. Such capability emerges from the orchestration of long-horizon planning, multi-hop retrieval, iterative tool use, dynamic reasoning and human-in-the-loop interaction. 

\begin{figure}[!t]
  \centering
    \includegraphics[width=1.0\linewidth]{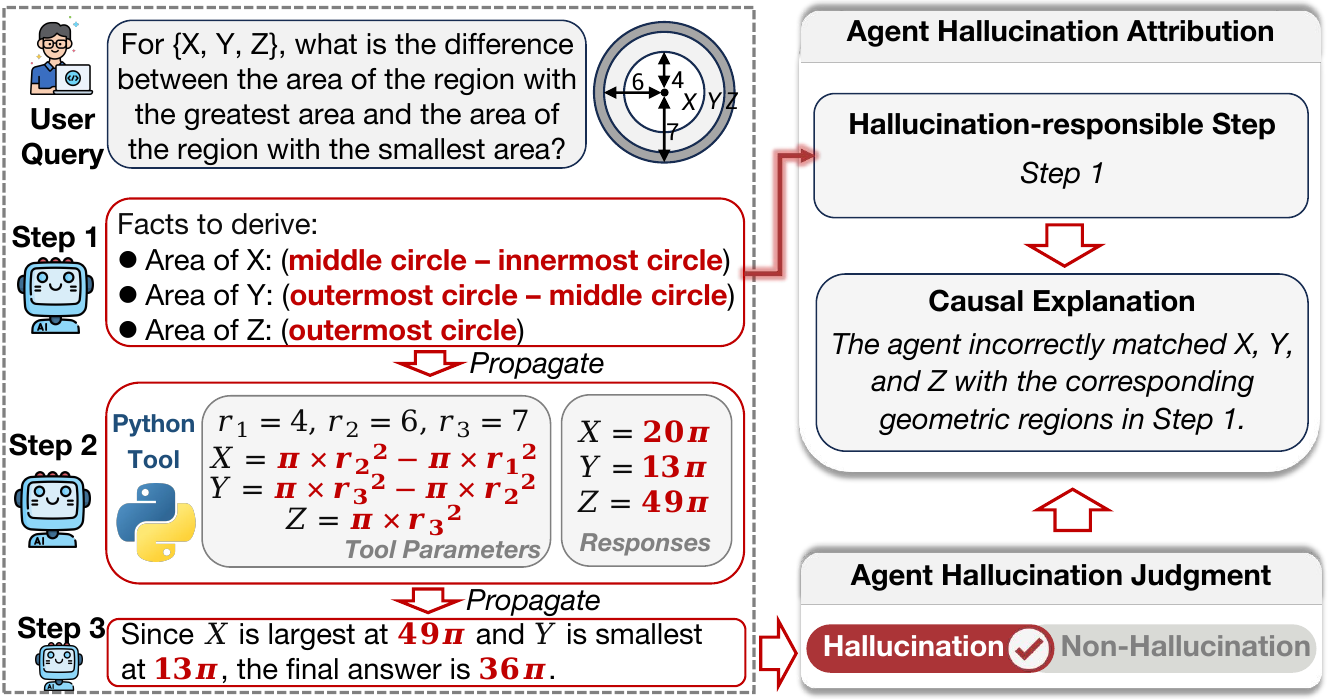}
    \caption{ Illustration of hallucination attribution in LLM-based agents. \textbf{Left:} A misdefinition of regions X, Y, Z in Step 1 propagates to the tool call and leads to the incorrect final answer. \textbf{Right:} Beyond binary judgment, hallucination attribution aims to identify a hallucination-responsible step and a causal explanation.
    }
    \label{fig:example_illus}
\end{figure}

However, hallucination, the generation of plausible yet non-factual content, remains a persistent issue in LLM-based systems. Unlike LLM hallucinations confined to single-turn responses~\cite{huang2025survey,ji2023survey}, agent-based hallucinations are amplified by the sequential nature of multi-step workflows, where intermediate errors propagate and ultimately degrade the final response~\cite{zhou2025guardian}. As shown in Figure~\ref{fig:example_illus} \textbf{Left}, a planning hallucination misdefines ``region X, Y, Z'', which propagates into downstream Python tool parameters and leads to an incorrect final answer. This underscores an urgent need for granular analyses to pinpoint the origin of the hallucination, especially in high-stakes agentic applications~\cite{huang2025survey}. 



\input{Table/dataset_comparison}

Current hallucination evaluations~\cite{bang2025hallulens,niu2024ragtruth,li2023halueval} primarily classify single-turn LLM responses as factual or hallucinated. While valuable, this binary paradigm fails to address concerns essential for building reliable agents: \textbf{where} and \textbf{why} hallucinations originate in agentic workflows. To fill this gap, we propose a novel research task of \textbf{automated hallucination attribution} for LLM-based agents. We define two key objectives: \textbf{(1) Hallucination-responsible Step Localization (Where):} identify the step responsible for the hallucinated result, (2) \textbf{Causal Explanation (Why):} provide an open-ended explanation of the underlying cause. As shown in Figure~\ref{fig:example_illus} \textbf{Right}, step attribution precisely identifies ``Step 1'' as the hallucination origin, while causal explanation provides a corresponding diagnostic analysis of ``Step 1 incorrectly matched X, Y, and Z with their regions''.

To support this task, we present AgentHallu, the first comprehensive benchmark tailored for automated hallucination attribution of multi-step agent trajectories. As shown in Table~\ref{dataset_comparison}, the key highlights of the AgentHallu dataset include: \textit{(1) Extensive Diversity.} We collect 693 trajectories from 7 popular agent frameworks with an average length of 7.6 steps. The dataset encompasses five distinct domains: world knowledge, science, math, general assistant, and tool use. \textit{(2) High-quality Control.} We implement a rigorous three-stage filtering criterion to exclude non-deceive failures, overly short sequences, and trivial cases lacking diagnostic depth, thereby ensuring the benchmark's difficulty. \textit{(3) Comprehensive Taxonomy.} We develop a hierarchical taxonomy of agent hallucinations via grounded theory~\cite{glaser2017discovery}, resulting in 5 primary categories (Planning, Retrieval, Reasoning, Human-Interaction, and Tool-Use) and 14 granular subcategories. \textit{(4) Multi-level Annotation.} AgentHallu includes binary labels for judgment. For attribution, it specifies hallucination-responsible steps and explains the underlying cause in plain language. All annotations are manually curated through a labor-intensive process.

Using the AgentHallu, we develop an attribution evaluation framework along two dimensions: step localization accuracy as a measure of responsible-step identification, and G-EVAL scores~\cite{liu2023g} for assessing the quality of open-ended explanations. Leveraging this framework, we evaluate 13 leading LLMs, including 5 proprietary and 8 open-source models. Empirical results reveal several critical findings: (1) The best-performing model, Gemini-2.5-Pro, achieves only 41.1\% accuracy in step localization, which drops to 11.6\% accuracy on tool-use hallucinations. (2) Step-by-Step prompting improves attribution via incremental processing, but at the cost of higher token usage. (3) Increasing trajectory steps $N_{\text{step}}$ poses a challenge to attribution, with GPT-5's accuracy dropping from 40.3\% ($N_{\text{step}} \le 5$) to 23.9\% ($N_{\text{step}} \ge 11$).




Overall, our contributions include: (i) A novel task of automated hallucination attribution in LLM-based agents to understand where and why hallucinations originate. (ii) A comprehensive benchmark comprising 693 high-quality trajectories with broad diversity, a systematic taxonomy and multi-level annotations. (iii) Evaluation of 13 leading LLMs, revealing their strengths and limitations under varying conditions, including hallucination categories, prompting methods, and trajectory steps.

%% file: Table/dataset_comparison.tex
\begin{table*}[t]
    \centering
    \tabcolsep=5pt
    \caption{Comparison of AgentHallu with existing hallucination detection datasets in terms of dataset statistics (sample size (\textbf{\#Samp.}) and trajectory steps (\textbf{\#Step})), hallucination categories (planning hallucination (\textbf{Planning}), retrieval hallucination (\textbf{Retrieval}), reasoning hallucination (\textbf{Reasoning}), human-interaction hallucination (\textbf{Human}), and tool-use hallucination (\textbf{Tool})), and task type (Hallucination \textbf{Judgment} and Hallucination \textbf{Attribution}).
}
       \vspace{-0.8em}
    \label{dataset_comparison}
    \resizebox{\linewidth}{!}{
\begin{tabular}{l|cc|ccccc|cc}
\toprule
\multirow{2}{*}{\textbf{Dataset}} & \multicolumn{2}{c|}{\textbf{Dataset Statistic}} & \multicolumn{5}{c|}{\textbf{Hallucination Category}}         & \multicolumn{2}{c}{\textbf{Task Type}} \\ \cline{2-10} 
                         & \textbf{\#Samp.}        & \textbf{\#Step}        & \textbf{Planning} & \textbf{Retrieval} & \textbf{Reasoning} & \textbf{Human} & \textbf{Tool} & \textbf{Judgment}     & \textbf{Attribution}    \\ \midrule \midrule
HaluEval~\cite{li2023halueval}                 & 35,000         & 1             & \crossmark     & \crossmark      & \checkmark      & \crossmark  & \crossmark & \checkmark         & \crossmark           \\
FELM~\cite{zhao2023felm}                     & 847            & 1             & \crossmark     & \crossmark      & \checkmark      & \crossmark  & \crossmark & \checkmark         & \crossmark           \\
SAC$^3$~\cite{zhang2023sac3}                     & 500            & 1             & \crossmark     & \crossmark      & \checkmark      & \crossmark  & \crossmark & \checkmark         & \crossmark           \\
FAVABench~\cite{mishra2024fine}                & 902            & 1             & \crossmark     & \crossmark      & \checkmark      & \crossmark  & \crossmark & \checkmark         & \crossmark           \\
RAGTruth~\cite{niu2024ragtruth}                 & 2,965          & 1             & \crossmark     & \checkmark      & \crossmark      & \crossmark  & \crossmark & \checkmark         & \crossmark           \\
ToolBH~\cite{zhang2024toolbehonest}                   & 700            & 1             & \crossmark     & \crossmark      & \crossmark      & \crossmark  & \checkmark & \checkmark         & \crossmark           \\
\midrule
\rowcolor{LightCyan}
\textbf{AgentHallu (Ours)}        & 693            & 7.6           & \checkmark     & \checkmark      & \checkmark      & \checkmark  & \checkmark & \checkmark         & \checkmark           \\ \bottomrule
\end{tabular}
    }
\end{table*}

%% file: Sec/2_related_work.tex
\section{Related Work}
\label{sec:realted_work}

\subsection{Hallucination Detection Benchmarks}
Hallucination detection aims to develop a framework or a model to automatically distinguish between hallucinated and factual content~\cite{ li2025hd,ravichander2025halogen,qin2025learning,zhang2025icr,zhang2025prompt}. As shown in Table~\ref{dataset_comparison}, a line of work assesses the model’s factuality reasoning over diverse domains such as world knowledge~\cite{li2023halueval, wei2024long,bang2025hallulens}, science~\cite{zhao2023felm}, and math~\cite{zhao2023felm}. Moreover, RAGTruth~\cite{niu2024ragtruth} demonstrates that popular LLMs continue to hallucinate across tasks even with retrieval-augmented generation. To diagnose tool-use hallucinations, ToolBH~\cite{zhang2024toolbehonest} collects 700 tool-call samples to perform solvability detection, solution planning, and missing-tool analysis. Different from prior works confined to binary judgment in single-turn responses, we introduce the first benchmark for automated hallucination attribution within multi-step agent trajectories.

\subsection{LLM-based Agents}
LLM-based agents~\cite{yao2022react,wang2024executable} have showcased extraordinary capabilities in automating tasks across various fields. This growing capability is largely driven by emergent behaviors that arise during chain-of-thought~\cite{wei2022chain}, in-context learning~\cite{brown2020language}, and instruction following~\cite{longpre2023flan}. To extend agents' capabilities beyond their internal knowledge, function calling~\cite{schick2023toolformer,patil2024gorilla,patilberkeley} has been proposed, enabling agents to interact with external tools and APIs in multi-step workflows.

Moreover, individual agents, each serving a specialized role, can be composed into multi-agent systems to solve complex tasks~\cite{hong2023metagpt, wang2024openhands}. Early multi-agents elicit stepwise reasoning through structured debate~\cite{liang2024encouraging} or role-play dialogue~\cite{li2023camel}. Recent works~\cite{fourney2024magentic, hu2025owl} introduce central orchestrators that assign tasks to specialized agents. Beyond inter-agent interaction, agents are motivated to proactively seek human feedback to improve their decision-making~\cite{feng2024large}. Despite this progress, hallucinations persist across operational stages in agent workflows~\cite{lin2025llm}, including planning, retrieval, reasoning, tool use, and human interaction, underscoring the need for reliability assessment.

%% file: Sec/3_method.tex
\section{Task Formulation}
In this section, we formulate the task of automated hallucination judgment and attribution. 

\begin{figure*}[!t]
  \centering
    \includegraphics[width=1.0\linewidth]{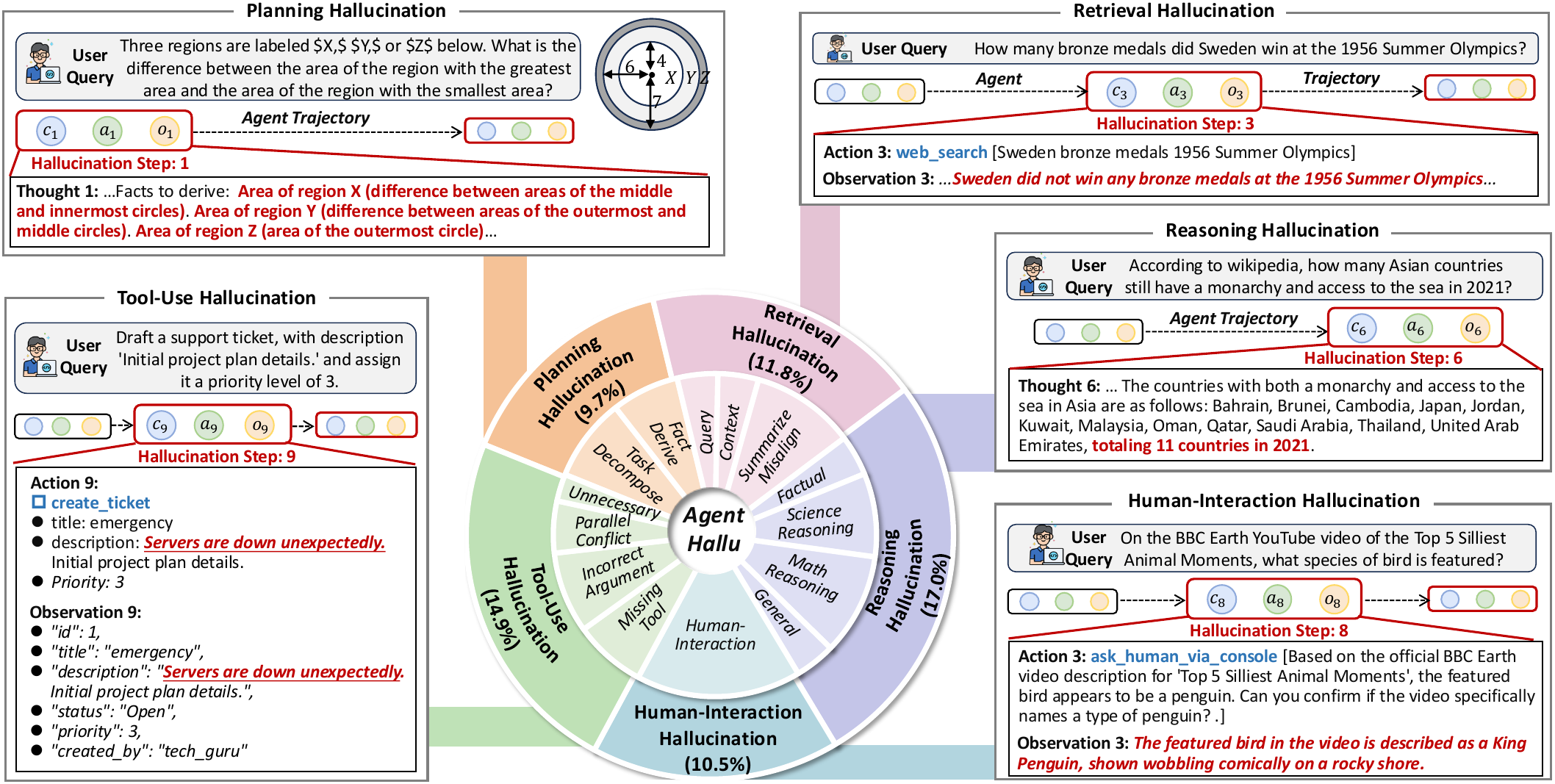}
    \caption{ Overview of hallucination taxonomy in AgentHallu. The dataset includes 5 hallucination categories and 14 subcategories, where each trajectory step interleaves a thought step, an action step, and an observation step.
    }
    \label{fig:dataset_ills}
\end{figure*}

\paragraph{Background.} 
LLM-based agents perform complex tasks with structured reasoning, where each interaction unit $u_t$ interleaves a thought step $c_t$, an action step $a_t$, and an observation step $o_t$. The trajectory $\tau$ can be written as:
\begin{equation}
    \tau = (u_1, u_2, \ldots, u_t),
\end{equation}
\begin{equation}
    u_t = (c_t, a_t, o_t),
\end{equation}
where $c_t$ denotes the internal reasoning state, $a_t$ specifies the invoked tool action, and $o_t$ captures feedback from the tool responses.
Distinct from prior analyses of non-hallucination failures~\cite{zhang2025agent,cemri2025multi,rahardja2025can}, we restrict our study to trajectories that yield coherent and seemingly plausible answers.

\paragraph{Hallucination Judgment Objective.}
We classify a trajectory as hallucinated by determining whether its produced answer diverges from the ground-truth solution corresponding to the task:
\begin{equation}
    \mathrm{is\_hallucination}(\tau) = \mathbbm{1}_{\left\{ y(\tau) \neq y^{\mathrm{gt}} \right\}},
\end{equation}
where $y(\tau)$ denotes the result of a trajectory $\tau$, $y^{\mathrm{gt}}$ denotes the task-specific ground-truth answer, and $\mathbbm{1}_{\left\{ \cdot \right\}}$ is the indicator function.

\paragraph{Hallucination Attribution Objective.}
Motivated by~\cite{zhang2025agent}, we identify a hallucinated step $u_{t}$ as the step whose correction is causally sufficient to transform an incorrect result into a correct one. Specifically, we replace $u_{t}$ with its correct counterpart and roll out the subsequent steps to obtain the counterfactual trajectory $\tau^{(t)}$. The set of hallucinated steps $\mathcal{H}(\tau)$ of a trajectory $\tau$ is then defined as:
\begin{equation}
    \mathcal{H}(\tau) = \{\, t \mid y(\tau) \neq y^{\mathrm{gt}} \land y(\tau^{(t)}) = y^{\mathrm{gt}} \,\}, 
\end{equation}
where $y(\tau^{(t)})$ denotes the result produced by the counterfactual trajectory $\tau^{(t)}$.
To address scenarios with multiple hallucinated steps, we follow a \textbf{causality-aligned principle} and treat the initial error as the primary source of hallucination. We thus define an objective to determine:
\begin{equation}
    t^{\star}= \arg\min_{\, t \in \mathcal{H}(\tau)} \; t .
\end{equation}
In this study, we address the problem of automatically identifying the step $t^{\star}$ and providing an associated open-ended explanation.

\section{AgentHallu Dataset}
\label{sec:dataset}
In this section, we first present an overview of our AgentHallu dataset in Sec.~\ref{overview}. Then, we detail the dataset development involving query collection in Sec.~\ref{query_collection}, trajectory construction in Sec.~\ref{trajectory_construction}, and hallucination annotation in Sec.~\ref{hallucination_annotation}.

\subsection{Overview}
\label{overview}
As shown in Table~\ref{dataset_comparison} and Figure~\ref{fig:dataset_ills}, AgentHallu comprises 693 annotated agent trajectories, including 443 hallucinated instances and 250 non-hallucinated instances. Each instance in AgentHallu includes the following entries: \textbf{(1) Query:} A real-world question from 8 datasets, covering domains spanning world knowledge, science, math, general assistant, and tool use. \textbf{(2) Trajectory:} A trajectory generated to address the query, with each step standardized into a triplet of thought, action, and observation. These trajectories are collected from 7 mainstream LLM-based agents. \textbf{(3) Annotation:} Multi-level annotations curated by human labelers, comprising a binary label, a hallucination-responsible step, and a causal explanation. Detailed dataset statistics are provided in Appendix~\ref{appendix_data_statis}.

\subsection{Query Collection}
\label{query_collection}
To ensure comprehensive coverage of factuality, we curate a diverse set of queries spanning five realistic domains, as detailed below.
\begin{itemize}[noitemsep,left=2pt, itemsep=0pt, topsep=0pt]
    \item \textit{World Knowledge}: We incorporate queries from the SimpleQA dataset~\cite{wei2024measuring}, spanning ten topics such as politics, art, and sports to represent general world knowledge.
    \item \textit{Science}: We include graduate-level scientific queries from the GPQA dataset~\cite{rein2024gpqa}, involving the disciplines of physics, chemistry, and biology.
    \item \textit{Math}: We filter out difficulty Level 1 and Level 2 questions from MATH-500~\cite{hendrycks2021measuring}, retaining the harder subset. To integrate frontier-level challenges, we also include questions from the American Invitational Mathematics Examination (AIME) 2024 and AIME 2025.
    \item \textit{General Assistant}: We include queries from the GAIA validation set~\cite{mialon2023gaia}, which provides diverse and realistic instructions reflecting general assistant use.
    \item \textit{Tool Use}: To mimic complex tool-use sequences in agentic workflows, we incorporate multi-turn and multi-step function-calling queries from BFCL V3~\cite{patilberkeley}.
\end{itemize}
To extend coverage toward cutting-edge human knowledge, we also include a small subset of questions from HLE~\cite{phan2025humanity}, spanning mathematics, humanities, and the natural sciences.

\subsection{Trajectory Construction}
\label{trajectory_construction}
Given the collected queries, we generate diverse and realistic trajectories by executing 7 widely used LLM-based agents (SmolAgents~\cite{roucher2025smolagents}, OpenDeepSearch~\cite{alzubi2025open}, OpenManus~\cite{openmanus2025}, OctoTools~\cite{lu2025octotools}, Magentic-One~\cite{fourney2024magentic}, OWL~\cite{hu2025owl}, and Function-calling Agents~\cite{patilberkeley}). Specifically, we partition queries from the four knowledge-intensive domains into six subsets and instantiate trajectories using the first six agents. In parallel, we utilize BFCL V3 to construct function-calling agent trajectories. These agents are primarily built on the GPT series (GPT-4o and GPT-4.1). Details on agent configuration are provided in Appendix~\ref{appendix_agent_descrip}. 

To enhance the robustness and quality of our benchmark, we apply a three-stage filtering criterion to the collected trajectories: 
\begin{itemize}[noitemsep,left=2pt, itemsep=0pt, topsep=0pt]
    \item \textit{(1) Exclude Failure Trajectories}: Since non-deceptive failures are easy to identify, we manually exclude trajectories that terminate without a task-completing response (e.g., early termination due to turn limits, token overflows, or tool-permission restrictions).
    \item \textit{(2) Exclude Short Trajectories}: Excessively short agent trajectories degrade into native LLM responses, lacking sufficient reasoning depth for step localization. Therefore, we exclude trajectories with only one or two valid steps.
    \item \textit{(3) Exclude Trivial Trajectories}: To select plausible and difficult samples, we retain trajectories with disagreement among LLM judges. Specifically, we use four independent LLMs (\texttt{GPT-5}, \texttt{Gemini-2.5-Pro}, \texttt{DeepSeek-V3.1}, and \texttt{Qwen3-32B}) to assign a binary label and a hallucination-responsible step for each trajectory. Then we exclude trajectories with full agreement across all four judges.
\end{itemize}

\subsection{Hallucination Annotation}
\label{hallucination_annotation}
Through the multi-step filtering described above, we retain 693 agent trajectories. To ensure consistent and reproducible annotations across heterogeneous agent systems, we establish both an empirically grounded hallucination taxonomy and a standardized annotation protocol.

\input{Table/category_defination}

\noindent\textbf{Empirically Grounded Taxonomy.}
To allow hallucination modes to emerge from empirical data, we apply grounded theory~\cite{glaser2017discovery} to analyze a pilot set of 140 trajectories sampled from seven agent frameworks. Specifically, we first perform open coding~\cite{khandkar2009open} on the trajectory data to label observed hallucinated behaviors. Then we apply constant comparative analysis to refine the boundaries between different hallucination types. By merging and linking relevant behaviors, we organize the open codes into a structured taxonomy of hallucination categories. The taxonomy is finally refined through discussion and review until consensus is reached. The resulting taxonomy is presented in Table~\ref{category_description}.

\input{Table/agreement}

\input{Table/model_performance}

\noindent\textbf{Standardized Annotation Protocol.}
We introduce a hallucination annotation protocol, which progresses from binary judgment to fine-grained attribution and taxonomy classification. To ensure annotation rigor, we employ ten graduate-level annotators with specialized expertise in AI to perform iterative labeling and refinement.
\begin{itemize}[noitemsep,left=2pt, itemsep=0pt, topsep=0pt]
    \item \textit{(1) Construction of Oracle-guided Reasoning Paths}. Considering that hallucination attribution often requires domain-specific expertise, we leverage LLMs to 
    construct detailed reasoning paths for question solving. Specifically, we condition the LLM on the question, the ground-truth answer, and, when available, dataset-provided solution annotations. Compared with question-only prompting, providing this additional information yields more faithful reasoning paths. To mitigate model-specific bias, each path is independently drafted by two different LLMs, GPT-5-Thinking and Gemini-2.5-Pro.   
    \item \textit{(2) Human Annotation}. Annotators first make a binary judgment of whether the agent trajectory is hallucinated by comparing it with the ground truth. For hallucinated cases, they further annotate the category, hallucination-responsible step, and causal explanation. To facilitate this process, LLMs are prompted with the reasoning path to generate attribution references, which are subsequently verified by annotators. Validation relies on two criteria: whether the candidate step introduces a factual error that directly distorts the outcome, or whether it propagates an error seeded in an earlier step. Upon detecting such propagation, annotators trace the error chain backward and reassign attribution to the root cause.
    \item \textit{(3) Consensus Resolution}. Inter-annotator agreement statistics are reported in Table~\ref{agreement}. For disagreements, annotations are resolved through collaborative discussion, requiring all annotators to be convinced by the final rationale. For agreed cases, we employ a cross-validation protocol in which each annotator reviews peer annotations to ensure adherence to shared standards. Any detected inconsistency triggers discussion and re-annotation until consensus is achieved.    
\end{itemize}

%% file: Table/category_defination.tex
\begin{table}[t!]
    \centering
    \caption{Taxonomy of agent hallucinations.}
       \vspace{-0.5em}
    \label{category_description}
    \resizebox{\linewidth}{!}{
    \tablestyle{3.0pt}{1.3}
    \begin{tabular}{ll|l}
\hline
\rowcolor{COLOR_MEAN}
\textbf{Category}                                                                            & \textbf{Sub-category} & \textbf{Description}                            \\ \hline \hline
\multirow{2}{*}{\textbf{\begin{tabular}[c]{@{}l@{}}Planning \\ Hallucination\end{tabular}}}  & Fact Derive           & Introduce nonexistent or misleading facts.      \\
                                                                                             & Task Decompose        & Produce task-misaligned subgoals.               \\ \hline
\multirow{3}{*}{\textbf{\begin{tabular}[c]{@{}l@{}}Retrieval \\ Hallucination\end{tabular}}} & Query Misalign        & Formulate inaccurate retrieval queries.         \\
                                                                                             & Context Misalign      & Retrieve factually incorrect context.           \\
                                                                                             & Summarize Misalign    & Misrepresent context via summarization.         \\ \hline
\multirow{4}{*}{\textbf{\begin{tabular}[c]{@{}l@{}}Reasoning\\ Hallucination\end{tabular}}}  & Factual Reasoning     & Incorrect factual inference over context.       \\
                                                                                             & Math Reasoning        & Incorrect math inference or computation.        \\
                                                                                             & Science Reasoning     & Incorrect science inference or computation.     \\
                                                                                             & General Reasoning     & Incorrect reasoning over general tasks.         \\ \hline
\multicolumn{2}{l|}{\textbf{Human-Interaction Hallucination}}                                                        & Incorrect messages propagated by the user.      \\ \hline
\multirow{4}{*}{\textbf{\begin{tabular}[c]{@{}l@{}}Tool-Use \\ Hallucination\end{tabular}}}  & Missing Tool          & Miss required tool invocation.                  \\
                                                                                             & Incorrect Argument    & Mis-specify arguments of invoked tools.         \\
                                                                                             & Parallel Conflict     & Trigger execution conflicts via parallel tools. \\
                                                                                             & Unnecessary Tool      & Invoke irrelevant or incorrect tools.           \\ \hline
\end{tabular}
    }
\end{table}

%% file: Table/agreement.tex
\begin{table}[t!]
    \centering
    \caption{Initial inter-annotator agreement on binary judgment (\textbf{Judgment}), categorization (\textbf{Category}), and hallucination-responsible step (\textbf{Step}). The results highlight the difficulty of manual hallucination attribution.}
       \vspace{-0.5em}
    \label{agreement}
    \resizebox{\linewidth}{!}{
    \tablestyle{3.0pt}{1.2}
    \begin{tabular}{l||ccccc|c}
\hline
\rowcolor{COLOR_MEAN}
Annotation   & World. & Science & Math & General & Tool & Overall \\ \hline \hline
Judgment   & 98.4        & 98.0         & 100.0      & 100.0         & 98.8      & 98.9        \\
Category & 83.2        & 74.3         & 81.2      & 74.4         & 92.2     & 81.9        \\
Step     & 80.4        & 72.5         & 76.5      & 69.2         & 85.4     & 77.9        \\ \hline
\end{tabular}
    }
\end{table}

%% file: Table/model_performance.tex
\begin{table*}[!t]
  \centering
  \caption{Performance (\%) of LLMs on AgentHallu under standard prompting, reporting hallucination judgment (F1/Recall/Acc) and hallucination attribution measured by step localization accuracy (Acc) and G-EVAL (GE).}
       \vspace{-0.8em}
  \label{model_comparision}
    \resizebox{1.0\linewidth}{!}{
    \tablestyle{5.0pt}{1.15}
   \begin{tabular}{lccccccccccccccc}
\toprule
\multicolumn{1}{l|}{\multirow{3}{*}{\textbf{Model Name}}} & \multicolumn{3}{c|}{\textbf{Judgment}}                             & \multicolumn{12}{c}{\textbf{Attribution}}                                                                                                                                                                      \\ \cline{2-16} 
\multicolumn{1}{l|}{}                                     & \multicolumn{3}{c|}{Overall}                                       & \multicolumn{2}{c}{Planning} & \multicolumn{2}{c}{Retrieval} & \multicolumn{2}{c}{Reasoning} & \multicolumn{2}{c}{Human}    & \multicolumn{2}{c|}{Tool-Use}                         & \multicolumn{2}{c}{Overall}  \\ \cline{2-16} 
\multicolumn{1}{l|}{}                                     & F1$\uparrow$            & Recall$\uparrow$        & \multicolumn{1}{c|}{Acc$\uparrow$}           & Acc           & GE           & Acc            & GE           & Acc            & GE           & Acc           & GE           & Acc           & \multicolumn{1}{c|}{GE}           & Acc$\uparrow$           & GE$\uparrow$           \\ \midrule \midrule
\multicolumn{1}{c|}{Random}                               & 48.5          & 49.6          & \multicolumn{1}{c|}{49.5}          & 9.6           & -            & 8.8            & -            & 10.3           & -            & 7.4           & -            & 7.2           & \multicolumn{1}{c|}{-}            & 8.7           & -            \\ \midrule \midrule
\rowcolor{COLOR_MEAN}
\multicolumn{16}{l}{Proprietary Large Language Models}                                                                                                                                                                                                                                                                                          \\
\multicolumn{1}{l|}{GPT-5}                                &  \cellcolor{LightCyan}\textbf{70.2} & \cellcolor{LightCyan}\textbf{73.2} & \multicolumn{1}{c|}{\cellcolor{LightCyan}\textbf{70.6}} & \cellcolor{LightCyan}\textbf{31.3} & \cellcolor{LightCyan}\textbf{2.3} & 26.8           & 1.7          & 57.6           & 3.0          & 39.7          & 2.6          & 4.9           & \multicolumn{1}{c|}{0.6}          & 32.7          & 2.0          \\
\multicolumn{1}{l|}{GPT-5-mini}                           & 65.0          & 67.3          & \multicolumn{1}{c|}{65.5}          & 29.9          & 2.1          & 28.1           & 1.5          & 53.4           & 2.8          & \cellcolor{LightCyan}\textbf{61.6} & \cellcolor{LightCyan}\textbf{3.3} & 3.9           & \multicolumn{1}{c|}{0.6}          & 35.0          & 2.0          \\
\multicolumn{1}{l|}{Gemini-2.5-Pro}                       & 64.6          & 64.2          & \multicolumn{1}{c|}{68.8}          & 25.4          & 2.2          & \cellcolor{LightCyan}\textbf{45.1}  & \cellcolor{LightCyan}\textbf{2.3} & \cellcolor{LightCyan}\textbf{64.4}  & \cellcolor{LightCyan}\textbf{3.2} & 50.7          & 2.8          & 14.6          & \multicolumn{1}{c|}{1.3}          & \cellcolor{LightCyan}\textbf{41.1} & \cellcolor{LightCyan}\textbf{2.4} \\
\multicolumn{1}{l|}{Gemini-2.5-Flash}                     & 65.3          & 65.4          & \multicolumn{1}{c|}{67.7}          & 20.9          & 2.1          & 42.7           & 2.1          & 54.2           & 2.7          & 43.8          & 2.6          & 15.5          & \multicolumn{1}{c|}{1.3}          & 36.3          & 2.1          \\
\multicolumn{1}{l|}{Claude-4.5-Sonnet}                    & 63.6          & 63.7          & \multicolumn{1}{c|}{66.1}          & 26.9          & \cellcolor{LightCyan}\textbf{2.3}          & 30.5           & 1.7          & 44.9           & 2.3          & 43.8          & 2.4          & \cellcolor{LightCyan}\textbf{19.4} & \multicolumn{1}{c|}{\cellcolor{LightCyan}\textbf{1.4}} & 33.4          & 2.0          \\ \midrule
\multicolumn{1}{l|}{Average}                              & 65.7          & 66.8          & \multicolumn{1}{c|}{67.7}          & 26.9          & 2.2          & 34.6           & 1.9          & 54.9           & 2.8          & 47.9          & 2.7          & 11.6          & \multicolumn{1}{c|}{1.0}          & 35.7          & 2.1          \\ \midrule \midrule
\rowcolor{COLOR_MEAN}
\multicolumn{16}{l}{Open-source Large Language Models}                                                                                                                                                                                                                                                                                          \\
\multicolumn{1}{l|}{DeepSeek-V3.1}                        & \cellcolor{LightCyan}\textbf{52.1} & 52.1          & \multicolumn{1}{c|}{\cellcolor{LightCyan}\textbf{55.4}} & \cellcolor{LightCyan}\textbf{14.9} & \cellcolor{LightCyan}\textbf{1.8} & 22.0           & \cellcolor{LightCyan}\textbf{1.6} & 27.1           & \cellcolor{LightCyan}\textbf{1.8} & 21.9          & 1.9          & 7.8           & \multicolumn{1}{c|}{\cellcolor{LightCyan}\textbf{0.7}} & 19.0          & \cellcolor{LightCyan}\textbf{1.5} \\
\multicolumn{1}{l|}{Qwen3-32B}                            & 51.8          & 53.0          & \multicolumn{1}{c|}{52.7}          & 7.5           & 1.5          & 19.5           & 1.1          & \cellcolor{LightCyan}\textbf{28.8}  & 1.7          & \cellcolor{LightCyan}\textbf{41.1} & \cellcolor{LightCyan}\textbf{2.1} & \cellcolor{LightCyan}\textbf{8.7}  & \multicolumn{1}{c|}{0.5}          & \cellcolor{LightCyan}\textbf{21.2} & 1.3          \\
\multicolumn{1}{l|}{Qwen3-8B}                             & 49.5          & 54.2          & \multicolumn{1}{c|}{49.5}          & 4.5           & 1.2          & \cellcolor{LightCyan}\textbf{28.1}  & 1.3          & 17.0           & 0.9          & 23.3          & 1.2          & 3.9           & \multicolumn{1}{c|}{0.2}          & 15.1          & 0.9          \\
\multicolumn{1}{l|}{Qwen2.5-72B}                          & 44.3          & 55.2          & \multicolumn{1}{c|}{46.0}          & 4.5           & 0.8          & 3.7            & 0.3          & 9.3            & 0.6          & 13.7          & 0.7          & 6.8           & \multicolumn{1}{c|}{0.5}          & 7.7           & 0.6          \\
\multicolumn{1}{l|}{Qwen2.5-32B}                          & 49.3          & \cellcolor{LightCyan}\textbf{56.3} & \multicolumn{1}{c|}{49.6}          & 4.5           & 1.1          & 1.2            & 0.5          & 15.3           & 0.9          & 12.3          & 0.8          & 6.8           & \multicolumn{1}{c|}{0.6}          & 8.6           & 0.8          \\
\multicolumn{1}{l|}{Qwen2.5-7B}                           & 43.9          & 51.1          & \multicolumn{1}{c|}{44.2}          & 0.0           & 0.6          & 6.1            & 0.5          & 5.9            & 0.4          & 13.7          & 0.8          & 6.8           & \multicolumn{1}{c|}{0.5} & 6.6           & 0.5          \\
\multicolumn{1}{l|}{Llama3.3-70B}                         & 40.4          & 54.2          & \multicolumn{1}{c|}{43.6}          & 10.5          & 0.9          & 4.9            & 0.4          & 6.8            & 0.3          & 5.5           & 0.3          & \cellcolor{LightCyan}\textbf{8.7}  & \multicolumn{1}{c|}{0.5}          & 7.2           & 0.4          \\
\multicolumn{1}{l|}{Llama3.1-8B}                          & 35.1          & 52.1          & \multicolumn{1}{c|}{40.3}          & 0.0           & 0.3          & 2.4            & 0.3          & 0.9            & 0.2          & 4.1           & 0.2          & 1.0           & \multicolumn{1}{c|}{0.1}          & 1.6           & 0.2          \\ \midrule
\multicolumn{1}{l|}{Average}                              & 45.8          & 53.5          & \multicolumn{1}{c|}{47.7}          & 5.8           & 1.0          & 11.0           & 0.8          & 13.9           & 0.8          & 17.0          & 1.0          & 6.3           & \multicolumn{1}{c|}{0.4}          & 10.9          & 0.8          \\ \bottomrule
\end{tabular}
}
\end{table*}

%% file: Sec/4_experiment.tex
\section{Experiments}
\label{sec:experi}

\subsection{Experimental Setup}

\noindent\textbf{Evaluated Models.}
We evaluate 13 frontier proprietary and open-source LLMs. The proprietary models include OpenAI’s GPT-5 and GPT-5-mini; Google’s Gemini-2.5-Pro and Gemini-2.5-Flash; and Anthropic’s Claude-4.5-Sonnet. The open-source models include DeepSeek’s DeepSeek-V3.1; Alibaba’s Qwen3 (8B/32B) and Qwen-2.5 (7B/32B/72B); and Meta’s Llama-3.3-70B and Llama-3.1-8B.

\noindent\textbf{Prompting Methods.}
We evaluate two baseline prompting methods: Standard Prompting and Step-by-Step Prompting. In Standard Prompting, the model receives the query and the full trajectory and is asked to perform hallucination judgment and attribution. In Step-by-Step Prompting, the model receives the query and the trajectory incrementally and determines at each step whether a hallucination occurs, terminating immediately upon detection. More details can be found in Appendix~\ref{appendix_evaluation_method}.

\noindent\textbf{Evaluated Metrics.}
For hallucination judgment, we evaluate binary classification performance using standard metrics, including macro-F1, macro-recall, and accuracy.
For hallucination attribution, we report step localization accuracy, defined as the proportion of hallucinated instances for which the model correctly identifies the responsible step.
In addition, we use G-EVAL~\cite{liu2023g} with GPT-5 as the evaluator to score explanation quality. More details can be found in Appendix~\ref{appendix_evaluation_metric}.

\subsection{Main Results}

\begin{figure*}[!t]
  \centering
    \includegraphics[width=1.0\linewidth]{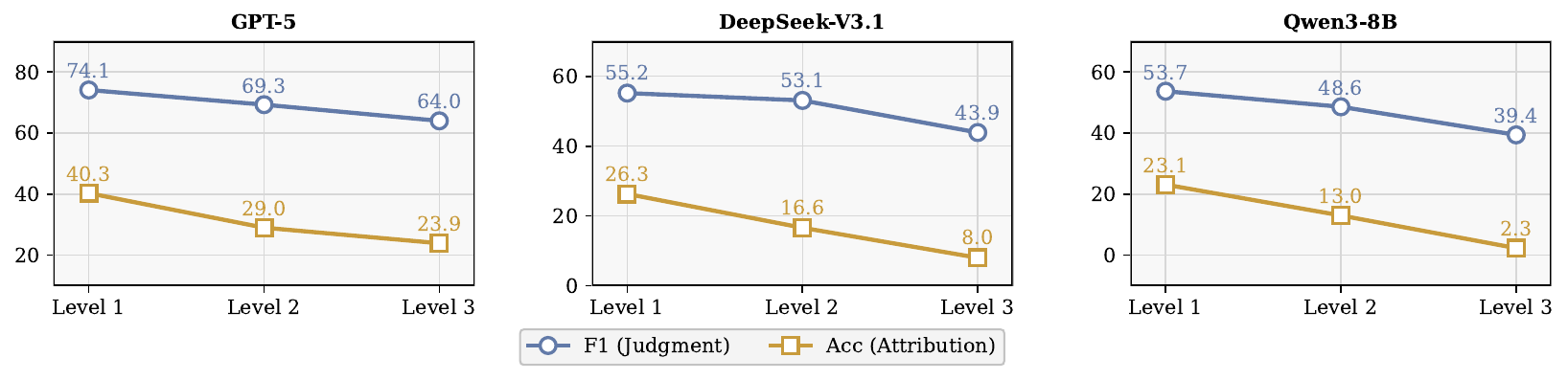}
    \vspace{-1.8em}
    \caption{Comparison of hallucination judgment and attribution performance across LLMs under varying trajectory steps $N_{\text{step}}$. Level 1 spans trajectories with $N_{\text{step}} \le 5$, Level 2 spans $6 \le N_{\text{step}} \le 10$, and Level 3 spans $N_{\text{step}} \ge 11$.
    }
    \label{fig:agent_length}
\end{figure*}

\paragraph{Comparison of Different LLMs.} Table~\ref{model_comparision} reports the main results of different LLMs on AgentHallu. Our key findings are summarized as follows:

\textit{(1) Challenges of Attribution}: A substantial performance gap remains between hallucination judgment and attribution tasks. While advances in proprietary models have boosted judgment performance to a peak F1 of 70.2\% for GPT-5, the more demanding attribution task reaches only 41.1\% localization accuracy and a 2.4 G-EVAL score for Gemini-2.5-Pro. These results indicate considerable room for attribution improvement and highlight the rigorous standards of this benchmark.

\textit{(2) Disparity between Proprietary and Open-source Models}: Open-source models achieve an average localization accuracy of 10.9\%, a level comparable to a random baseline and substantially below 35.9\% of proprietary models. Even the strongest open-source model, DeepSeek-V3.1, attains only 19.2\% localization accuracy. This performance gap may be attributed to the limited reasoning capabilities of open-source models.

\textit{(3) Category-level Analysis}:
Attribution accuracy varies substantially across hallucination categories. Reasoning hallucinations are comparatively easier to localize, with Gemini-2.5-Pro reaching 64.4\% accuracy, whereas tool-induced hallucinations remain consistently the most difficult across all model families. This may be attributed to the challenge of verifying environmental state within action–observation loops, rather than purely linguistic factual errors. Further analysis on subcategories is provided in Appendix~\ref{appendix_subcategory_analysis}.

\input{Table/method_comparison}

\paragraph{Comparison of Different Prompting Methods.}
We compare two prompting methods for hallucination judgment and attribution in Table~\ref{method_comparision}.
For hallucination judgment, the standard prompt remains consistently competitive and slightly outperforms the step-by-step variant, suggesting that binary decisions benefit from aggregating evidence over the full trajectory. In contrast, the step-by-step method substantially improves attribution, raising accuracy from 24.3\% to 36.6\% on average by incrementally processing context to enable more focused step localization. However, this method comes at a clear efficiency trade-off, increasing the average input token cost from 6,312 to 17,514 due to the additional decisions with multi-turn prompting.

\paragraph{Performance across Varying Trajectory Steps.}
To further examine the effect of trajectory steps on hallucination diagnosis, we partition the trajectory logs from AgentHallu into three levels based on the number of steps $N_{\text{step}}$. Level 1 includes trajectories with $N_{\text{step}} \le 5$, Level 2 covers $6 \le N_{\text{step}} \le 10$, and Level 3 contains $N_{\text{step}} \ge 11$, resulting in 278, 274, and 141 samples, respectively. Both judgment and attribution performances for three LLMs across these levels are presented in Figure~\ref{fig:agent_length}. The results show a consistent degradation in both tasks as trajectory length increases. Notably, attribution accuracy drops significantly on average, from 29.9\% at Level 1 to 11.4\% at Level 3, suggesting that the accumulation of distracting context can effectively obscure the hallucination-responsible step.

\begin{figure*}[!ht]
  \centering
    \includegraphics[width=1.0\linewidth]{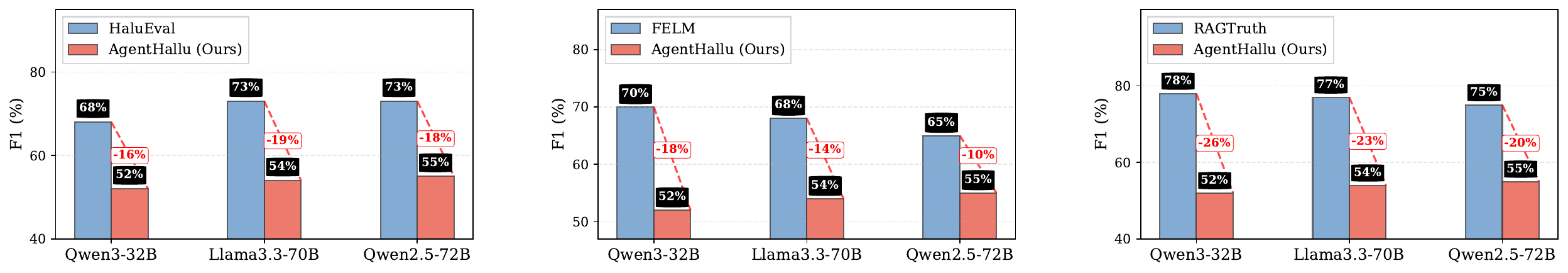}
    \vspace{-1.8em}
    \caption{ Comparison of hallucination judgment F1 (\%) on AgentHallu against existing hallucination detection datasets across multiple LLMs.
    }
    \label{fig:hallu_dataset}
\end{figure*}

\input{Table/evaluation_consistency}

\paragraph{Human Evaluation on Explanations.}
To examine alignment between causal explanation evaluation and human preference, we conduct a user study on 100 curated trajectory–explanation pairs from AgentHallu. The pairs are uniformly sampled across five hallucination categories and span outputs from five models, including GPT-5, Gemini-2.5-Pro, DeepSeek-V3.1, Qwen3-32B, and Llama3.3-70B. Three annotators with AI expertise independently rate each pair on a five-point scale. In Table~\ref{evaluation_consistency}, LLM-based evaluator, G-EVAL, align more closely with human judgments than Rouge-L~\cite{lin2004rouge} and BERTScore~\cite{zhang2019bertscore}. In particular, GPT-5-based G-EVAL achieves 0.86 Spearman and 0.76 Kendall-Tau, indicating reliable assessment of hallucination explanations.

\subsection{Experimental Analysis}

\paragraph{Comparison against Hallucination Detection Datasets.} To underscore the challenge of AgentHallu, we compare it against three existing hallucination detection datasets, HaluEval~\cite{li2023halueval}, FELM~\cite{zhao2023felm}, and RAGTruth~\cite{niu2024ragtruth}, using three advanced LLMs. As shown in Figure~\ref{fig:hallu_dataset}, all models consistently yield substantially lower performance on AgentHallu than on prior datasets, with an average degradation of about 18.2\% binary F1. This consistent difficulty stems from the long-horizon nature of multi-step trajectories and the broader coverage of hallucination categories in AgentHallu.

\begin{figure}[!ht]
  \centering
    \includegraphics[width=1.0\linewidth]{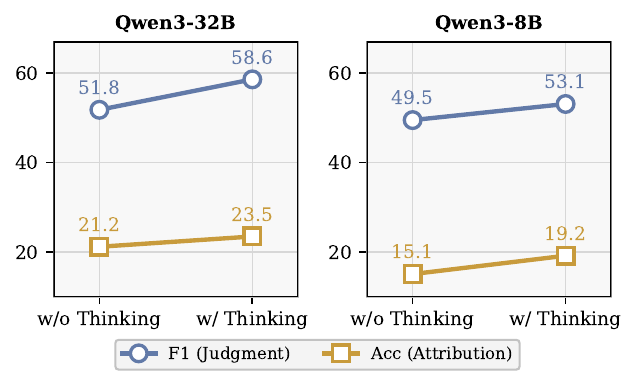}
    \vspace{-1.6em}
    \caption{ The performance of judgment and attribution with and without thinking mode on Qwen3.
    }
    \label{fig:thinking_mode}
\end{figure}

\paragraph{Effect of Thinking Mode.}
We further study the effect of enabling thinking mode on automated hallucination judgment and attribution. Figure~\ref{fig:thinking_mode} shows consistent gains for both Qwen3 variants when thinking is enabled. For Qwen3-32B, judgment F1 improves from 51.8 to 58.6, while attribution accuracy increases from 21.2 to 23.5. The improvements are primarily attributable to enhanced self-verification under thinking mode, which better distinguishes plausible yet incorrect claims.

\begin{figure}[!ht]
  \centering
    \includegraphics[width=1.0\linewidth]{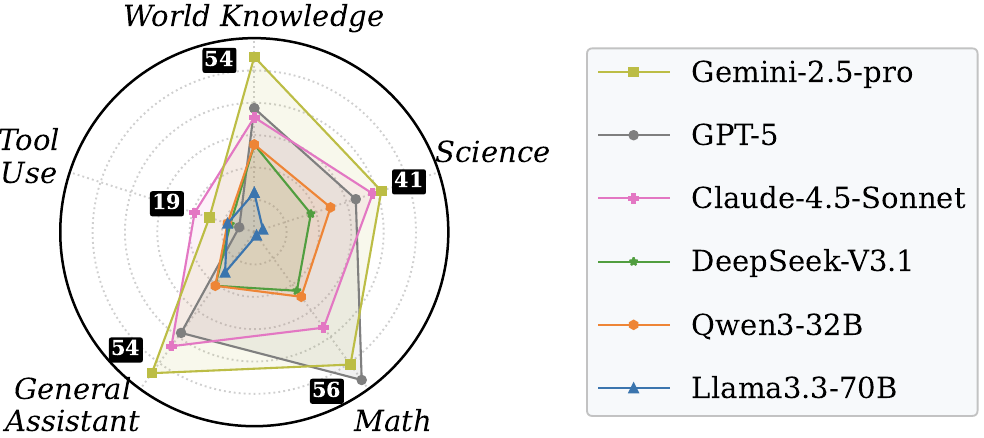}
    \vspace{-1.6em}
    \caption{ Step localization accuracy of six evaluated LLMs across five domains.
    }
    \label{fig:agent_domain}
\end{figure}

\paragraph{Performance across Domains.}
We finally report step localization accuracy across five domains in Figure~\ref{fig:agent_domain}. The results show that attribution remains challenging across all models and varies substantially by domain. For knowledge-intensive queries, performance peaks on Math at 56\% accuracy but drops notably on Science to 41\% accuracy. Tool Use is consistently the hardest, suggesting current LLMs are challenging to precisely track environment states under sequential tool interactions.

%% file: Table/method_comparison.tex
\begin{table}[!ht]
  \centering
  \caption{Comparison of prompting methods in terms of hallucination judgment (Judg.) F1, attribution (Attr.) step localization accuracy, and inference token cost.}
      \vspace{-0.5em}
  \label{method_comparision}
    \resizebox{1.0\linewidth}{!}{
    \tablestyle{5.0pt}{1.1}
   \begin{tabular}{c|c|c|c|c}
\toprule
\rowcolor{COLOR_MEAN} 
\cellcolor{COLOR_MEAN}                                                                                 & \cellcolor{COLOR_MEAN}                                                                                      & \textbf{Judg.} & \textbf{Attr.} & \textbf{Efficiency} \\ \cline{3-5} 
\rowcolor{COLOR_MEAN} 
\multirow{-2}{*}{\cellcolor{COLOR_MEAN}\textbf{\begin{tabular}[c]{@{}c@{}}Model \\ Name\end{tabular}}} & \multirow{-2}{*}{\cellcolor{COLOR_MEAN}\textbf{\begin{tabular}[c]{@{}c@{}}Prompting\\ Method\end{tabular}}} & \textbf{F1}$\uparrow$       & \textbf{Acc}$\uparrow$         & \textbf{Token Cost} \\ \midrule \midrule
                                                                                                         & Standard                                                                                                      & \textbf{70.2}     & 32.7                 & 7,426               \\
\multirow{-2}{*}{GPT-5}                                                                                  & Step-by-Step                                                                                                  & 68.5              & \textbf{42.7}        & \textbf{25,454}     \\ \midrule
                                                                                                         & Standard                                                                                                      & \textbf{52.1}     & 19.0                 & 6,494               \\
\multirow{-2}{*}{\begin{tabular}[c]{@{}c@{}}DeepSeek\\ -V3.1\end{tabular}}                               & Step-by-Step                                                                                                  & 51.1              & \textbf{35.9}        & \textbf{11,457}     \\ \midrule
                                                                                                         & Standard                                                                                                      & 51.8              & 21.2                 & 5,017               \\
\multirow{-2}{*}{\begin{tabular}[c]{@{}c@{}}Qwen3\\ -32B\end{tabular}}                                   & Step-by-Step                                                                                                  & \textbf{52.2}     & \textbf{31.2}        & \textbf{15,630}     \\ \bottomrule
\end{tabular}
}
\end{table}

%% file: Table/evaluation_consistency.tex
\begin{table}[ht!]
    \centering
    \caption{Spearman and Kendall-Tau correlations between different metrics and human annotations.}
    \vspace{-0.5em}
    \label{evaluation_consistency}
    \resizebox{\linewidth}{!}{
    \tablestyle{3.0pt}{1.1}
    \begin{tabular}{l||cc}
\toprule
\rowcolor{COLOR_MEAN} 
\textbf{Evaluation Metric} & \textbf{Spearman} & \textbf{Kendall-Tau} \\ \midrule
Rouge-L           &0.44           &0.34              \\
BERTScore         &0.32           &0.24              \\
G-EVAL-Qwen3-32B  &0.78           &0.62              \\
G-EVAL-GPT5       &\textbf{0.86}           &\textbf{0.76}              \\ \bottomrule
\end{tabular}
    }
\end{table}

%% file: Sec/5_conclusion.tex
\section{Conclusion}
\label{sec:conclu}
In this paper, we propose a novel task of automated hallucination attribution of LLM-based agents, aiming to identify the step where the initial hallucination originates and explain why it occurs.
To advance this task, we present AgentHallu, a comprehensive benchmark comprising 693 high-quality trajectories featuring: (1) extensive diversity spanning 7 agent frameworks and 5 domains, (2) systematic coverage of 5 hallucination categories and 14 subcategories, and (3) multi-level human annotations of binary labels, hallucination-responsible steps and causal explanations.
Evaluations on 13 leading LLMs highlight significant challenges, with performance varying across hallucination categories, prompting methods and trajectory lengths.

%% file: Sec/6_limitation.tex
\section*{Limitations}
While our AgentHallu marks a critical advancement in automated hallucination attribution for LLM-based agents, it is important to recognize several limitations. First, although AgentHallu spans 5 primary categories and 14 subcategories, it remains challenging to fully anticipate and represent emerging hallucination patterns as agent frameworks, tool ecosystems, and interaction protocols rapidly evolve. Therefore, the dataset should be continuously expanded to keep pace with new agent capabilities. Second, AgentHallu primarily targets text-based trajectories and does not consider multimodal agent settings grounded in images, audio, or other modalities. Given the growing adoption of multimodal agents, future work should explore extending the attribution framework to encompass these broader multimodal interactions.


\section*{Ethical Considerations}
AgentHallu is strictly free of personally identifiable information and offensive content. The benchmark is exclusively sourced from publicly accessible datasets and repositories, as well agent trajectories generated under controlled settings, explicitly avoiding sensitive or restricted data sources. Designed for academic research, AgentHallu focuses on enhancing the reliability of autonomous agents. Through adherence to strict data integrity protocols and ethical standards, AgentHallu establishes a responsible foundation for the automated attribution of agent hallucinations.

%% file: Sec/7_appendix.tex


\section{More Details on AgentHallu}
\subsection{Dataset and Code Release}
\label{appendix_data_code_realease}
The dataset and code are distributed under the Creative Commons Attribution-NonCommercial-ShareAlike 4.0 International License (CC BY-NC-SA 4.0). This license permits sharing and adaptation with appropriate attribution for non-commercial use, provided that derivative works are distributed under the same terms.

\subsection{Dataset Statistics}
\label{appendix_data_statis}

\subsubsection{Category Statistics}
AgentHallu includes 693 trajectories across one non-hallucination category and five hallucination categories and fourteen subcategories, with distribution statistics presented in Table~\ref{tab:category_ratio}. As shown in Table~\ref{tab:category_ratio}, the category distribution is intentionally balanced to ensure comprehensive and equitable coverage of diverse hallucination types. 

\input{Table/category_ratio}

\subsubsection{Word Cloud}
\begin{figure}[!t]
  \centering
    \includegraphics[width=1.0\linewidth]{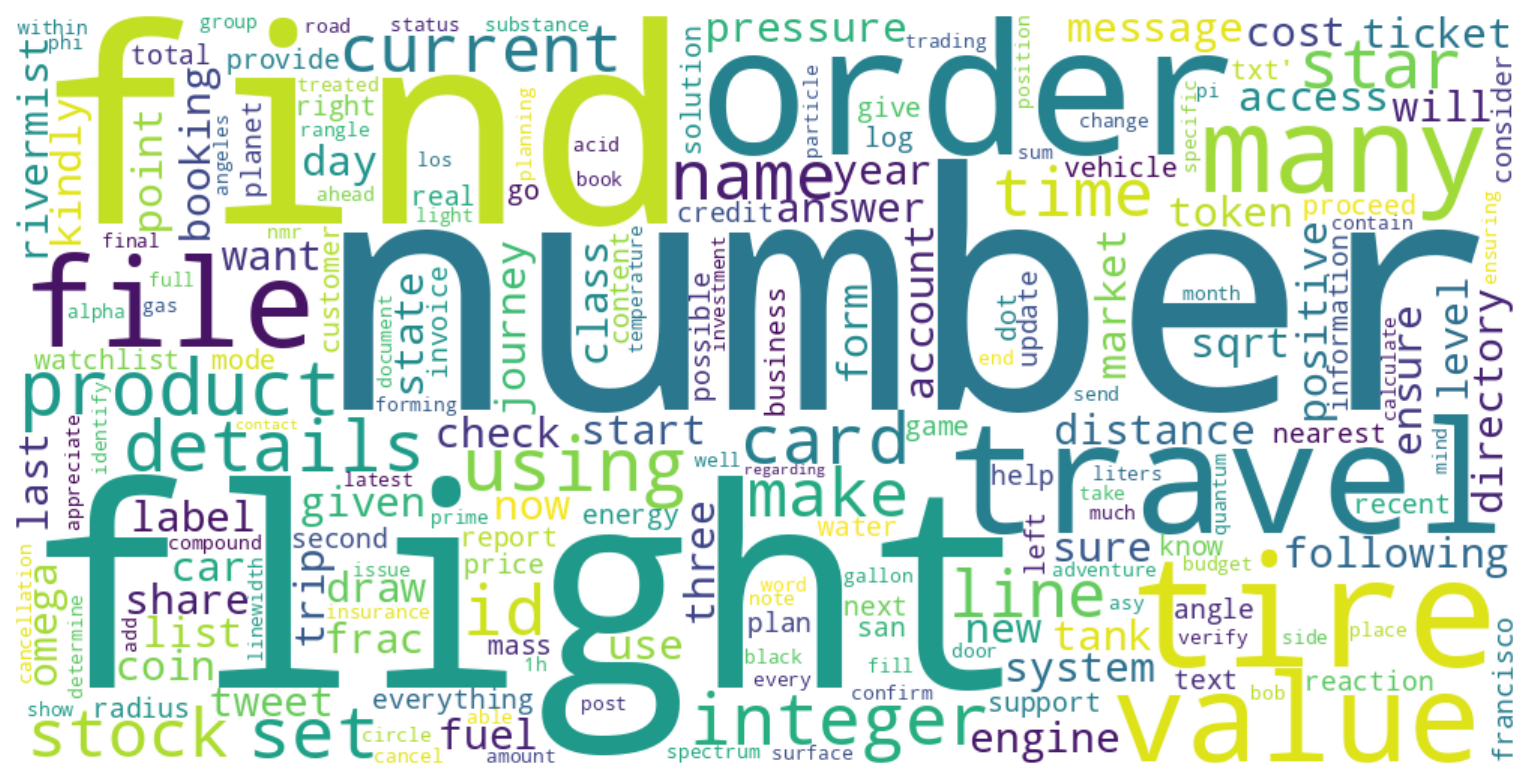}
    \caption{ Word cloud of queries in AgentHallu dataset.
    }
    \label{appendix_word_cloud}
\end{figure}

The query distribution in AgentHallu is visualized as a word cloud in Figure~\ref{appendix_word_cloud}. Prevalent high-frequency terms highlight recurring linguistic patterns in agent queries, including tool invocations and information-seeking requests. The breadth of salient keywords indicates substantial topical diversity, suggesting that the benchmark covers a wide range of realistic user intents rather than a narrow set of prompt templates.

\subsubsection{Trajectory Distribution}
\begin{figure}[!ht]
  \centering
    \includegraphics[width=1.0\linewidth]{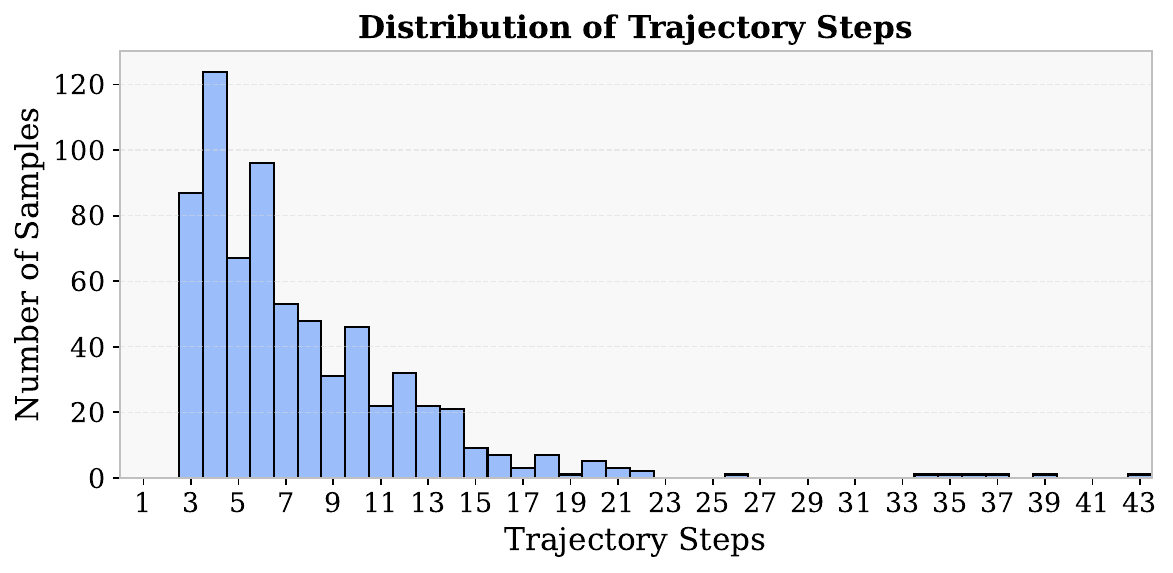}
    \caption{ Distribution of trajectory lengths measured by the number of steps per trajectory.
    }
    \label{appendix_tragectory_step}
\end{figure}

We present the trajectory length distribution of AgentHallu, measured by the number of steps per trajectory in Figure~\ref{appendix_tragectory_step}. As shown in Figure~\ref{appendix_tragectory_step}, AgentHallu excludes overly short trajectories with one or two steps. The longest trajectory contains 43 steps, and lengths are broadly distributed across step counts, indicating substantial diversity in interaction depth and long-horizon reasoning.

\input{Table/agent_dataset_distribution}

\subsection{Agent Configuration}

\subsubsection{Agent Description}
\label{appendix_agent_descrip}
We instantiate agent trajectories using seven representative LLM-based agents that span diverse reasoning paradigms and interaction patterns. We briefly summarize each framework below.
\begin{itemize}[noitemsep,left=2pt, itemsep=0pt, topsep=0pt]
    \item \textit{SmolAgents}~\cite{roucher2025smolagents}: SmolAgents is a lightweight agent framework that supports both CodeAct-style~\cite{wang2024executable} and ReAct-style~\cite{yao2022react} agents, and we configure both agent types within our method.
    \item \textit{OpenDeepSearch}~\cite{alzubi2025open}: Built on SmolAgents as the reasoning agent, OpenDeepSearch integrates an advanced search tool that leverages an embedded LLM to refine retrieved context. We also configure both CodeAct-style and ReAct-style agent variants.
    \item \textit{OpenManus}~\cite{openmanus2025}: OpenManus extends the planner–toolcaller architecture by incorporating explicit human-in-the-loop interactions, enabling the agent to solicit user input and integrate feedback during task execution.
    \item \textit{OctoTools}~\cite{lu2025octotools}: OctoTools provides over ten standardized tool cards that encapsulate diverse functionalities, enabling efficient multi-tool workflows for complex computational tasks.
    \item \textit{Magentic-One}~\cite{fourney2024magentic}: Magnetic-One employs a coordinator agent that collaborates with four specialized agents: a WebSurfer agent to browse the web, a FileSurfer agent to handle files, a Coder agent to write code, and a Computer Terminal agent to execute code.
    \item \textit{OWL}~\cite{hu2025owl}: OWL includes a workforce-oriented framework with a Planner for task decomposition, a Coordinator for subtask management, and specialized Workers capable of domain-specific tool invocation.
    \item \textit{Function-calling Agent}~\cite{patilberkeley}: A function-calling agent conditions an LLM on a set of tool or API specifications. The agent then emits a structured function call that selects the appropriate function and fills in its arguments. The tool output is fed back to the agent, enabling multi-turn execution and iterative reasoning.
\end{itemize}

\input{Table/agent_model_distribution}

\subsubsection{Agent Distribution across Datasets}
We present the distribution of query dataset sources across agent frameworks in AgentHallu, as shown in Table~\ref{agent_dataset_distribution}. The six general agent frameworks contribute trajectories across all seven knowledge-intensive datasets, yielding broad and relatively balanced coverage. In contrast, the function-calling agent is used exclusively for BFCL V3 queries to assess tool selection and argument filling behavior. Overall, we obtain 693 trajectories spanning heterogeneous agent designs and data sources, suggesting that AgentHallu captures diverse execution patterns and task contexts rather than artifacts of a specific agent implementation.

\input{Table/model_version}

\subsubsection{Agent Distribution across Models}
To enrich behavioral diversity and mitigate backbone-specific bias, we instantiate the six general agent frameworks with five LLM backbones (\texttt{GPT-5}, \texttt{GPT-4.1}, \texttt{GPT-4o}, \texttt{Claude-3.7-Sonnet}, and \texttt{Qwen2.5-Coder-32B}). For BFCL V3 queries, we incorporate trajectories generated by function-calling agents based on \texttt{GPT-4.1}, \texttt{Qwen3-32B}, and \texttt{Llama-3.3-70B}. We summarize the backbone composition per framework in Table~\ref{tab:agent_model_distribution}. The results reflect that this heterogeneous backbone mixture broadens AgentHallu’s behavioral diversity across frameworks and reduces reliance on any single model family, making the benchmark more representative for attribution evaluation.

\subsection{Source Dataset Licenses}
The licenses for the source query datasets used in this paper summarized are as follows:
\begin{itemize}[noitemsep,left=2pt, itemsep=0pt, topsep=0pt]
    \item \textit{SimpleQA}~\cite{wei2024measuring}: MIT License.
    \item \textit{GPQA}~\cite{rein2024gpqa}: MIT License.
    \item \textit{MATH-500}~\cite{hendrycks2021measuring}: MIT License. 
    \item \textit{AIME 2024 and AIME 2025}: MIT License. 
    \item \textit{GAIA}~\cite{mialon2023gaia}: The dataset does not specify an explicit license.
    \item \textit{BFCL V3}~\cite{patilberkeley}: Apache-2.0 License.
    \item \textit{HLE}~\cite{phan2025humanity}: MIT License.
\end{itemize}

\section{More Details on Evaluation}

\subsection{Model Configurations}
Table~\ref{model_version} summarizes the configurations of the LLM backbones used to generate agent trajectories and hallucination evaluation. For trajectory generation, we adopt the default LLM settings provided by each agent framework. For hallucination evaluation, to ensure fair comparisons, we fix the sampling hyperparameters by setting “do\_sample = False” or “Temperature = 0” to guarantee deterministic outputs, with the maximum output length set to 1024 tokens. All experiments are performed on eight NVIDIA GeForce A100 GPUs with PyTorch and are fully reproducible.

\begin{algorithm}[htb]
\setlength{\lineskip}{0.1em}
\setlength{\lineskiplimit}{0.1em}
\caption{Standard Prompting}
\label{alg:standard_prompting}
\begin{algorithmic}[1]
\Require Query $Q$, trajectory $\tau = (u_1, \ldots, u_n)$, llm evaluator $\mathcal{M}_{\theta}$
\Ensure Hallucination label $h\in\{0,1\}$, responsible step $s^*$, causal explanation $e^*$

\State $h \gets 0$; $s^* \gets \varnothing$; $e^* \gets \varnothing$
\State $(h, s^*, e^*) \gets \mathcal{M}_{\theta}(Q, \tau)$
\State \Return $h, s^*, e^*$ \Comment{$h=0$ indicates non-hallucination}
\end{algorithmic}
\end{algorithm}

\subsection{Prompting Method}
\label{appendix_evaluation_method}
We provide more details on two baseline prompting methods, described as follows:
\begin{itemize}[noitemsep,left=2pt, itemsep=0pt, topsep=0pt]
\item \textit{Standard Prompting Method:} Standard prompting feeds the query and the complete trajectory to an evaluator model in a single pass. The model is instructed to determine whether the trajectory contains a hallucination and, if hallucinated, to identify the earliest responsible step and provide a causal explanation linking that step. The algorithm of the standard prompting is summarized in Algorithm~\ref{alg:standard_prompting}.

\item \textit{Step-by-Step Prompting Method:} 
Step-by-Step prompting evaluates the trajectory in an incremental manner. It presents the query and trajectory prefixes step by step, and at each step determines whether the current prefix already contains a hallucination. The procedure terminates upon the first hallucination identification, assigning that step as the responsible step,  along with the causal explanation provided by the evaluator. The algorithm of the Step-by-Step prompting is summarized in Algorithm~\ref{alg:step_by_step_prompting}.
\end{itemize}

\begin{algorithm}[htb]
\setlength{\lineskip}{0.1em}
\setlength{\lineskiplimit}{0.1em}
\caption{Step-by-Step Prompting}
\label{alg:step_by_step_prompting}
\begin{algorithmic}[1]
\Require Query $Q$, trajectory $\tau = (u_1, \ldots, u_n)$, llm evaluator $\mathcal{M}_{\theta}$
\Ensure Hallucination label $h\in\{0,1\}$, responsible step $s^*$, causal explanation $e^*$

\State $h \gets 0$; $s^* \gets \varnothing$; $e^* \gets \varnothing$
\For{$i \in \{1, 2, \dots, n\}$}
    \State $\tau_{\le i} \gets (u_1, \ldots, u_i)$
    \State $(h_i, e_i) \gets \mathcal{M}_{\theta}(Q, \tau_{\le i})$
    \If{$h_i \!=\! 1$}
        \State $h \gets 1$
        \State $s^* \gets i$
        \State $e^* \gets e_i$ 
        \State \Return $h, s^*, e^*$
    \EndIf
\EndFor
\State \Return $h, s^*, e^*$ \Comment{$h=0$ indicates non-hallucination}
\end{algorithmic}
\end{algorithm}

\subsection{Evaluated Metric}
\label{appendix_evaluation_metric}
\paragraph{Hallucination Judgment.} 
For hallucination judgment, we adopt the widely used macro-F1 metric, which balances precision and recall through a harmonic mean. The macro-F1 score is computed as follows:

\begin{equation}
   \mathrm{macro\text{-}F1} = \frac{1}{K} \sum_{k=1}^{K} \frac{2\times \mathrm{Precision}_k\times \mathrm{Recall}_k}{\mathrm{Precision}_k+\mathrm{Recall}_k}.
\end{equation}
In this context, $K$ denotes the number of classes, and we set $K=2$ for binary classification. $\mathrm{Precision}_k$ denotes the class-$k$ precision, defined as the proportion of samples predicted as class $k$ that truly belong to class $k$:
\begin{equation}
      \mathrm{Precision}_k = \frac{TP_k}{TP_k + FP_k}.
\end{equation}
$\mathrm{Recall}_k$ is the recall for class $k$, defined as the proportion of samples from class $k$ that are correctly identified:
\begin{equation}
     \mathrm{Recall}_k = \frac{TP_k}{TP_k + FN_k}.
\end{equation}
Beyond the F1 score, we also include macro-recall and accuracy. Macro-recall is defined as follows:
\begin{equation}
   \mathrm{macro\text{-}Recall} = \frac{1}{K} \sum_{k=1}^{K} \mathrm{Recall}_k.
\end{equation}
The accuracy score is defined as follows:
\begin{equation}
   \mathrm{Accuray} = \frac{N_{\mathrm{correct}}}{N_{\mathrm{total}}},
\end{equation}
where $N_{\mathrm{correct}}$ is the number of correctly classified samples, and $N_{\mathrm{total}}$ is the total number of evaluated samples.

\paragraph{Hallucination Attribution.} 
Since a decisive hallucination step is well-defined only for hallucinated outputs, we compute attribution metrics on the subset of hallucinated samples. This restriction prevents non-hallucinated cases from dominating the score and keeps the metric aligned with responsible-step localization. For step localization, We report localization accuracy, defined as the proportion of samples for which the predicted step matches the ground-truth hallucination annotation. The localization accuracy is computed as follows:
\begin{equation}
\mathrm{Acc}_{\text{step}}
= \frac{1}{|\mathcal{H}_{hal}|}
\sum_{i \in \mathcal{H}_{hal}} \mathbbm{1}\!\left\{\hat{t}_{i} = t^{\ast}_{i}\right\},
\end{equation}
where $\mathcal{H}_{hal}$ denotes the subset of hallucinated samples, $t^{\ast}_{i}$ denotes the ground-truth hallucination-responsible step for sample $i$, $\hat{t}_{i}$ denotes the step predicted by the model, and $\mathbbm{1} {\left\{ \cdot \right\}}$ is the indicator function.

To further assess the quality of the causal explanations produced by each model, we adopt G-EVAL~\cite{liu2023g} and use GPT-5 as the evaluator. For each instance $i$, GPT-5 assigns an ordinal score $s_i \in {1,2,3,4,5}$. The score is determined by a fixed rubric that measures the explanation accuracy, guided by the human-annotated explanation and the trajectory. The full prompt template and scoring rubric are provided in Appendix~\ref{appendix_prompt_geval}.


\section{Additional Experiments}

\subsection{Model Bias Analysis in Explanation Evaluation.}
\begin{figure}[!ht]
  \centering
    \includegraphics[width=1.0\linewidth]{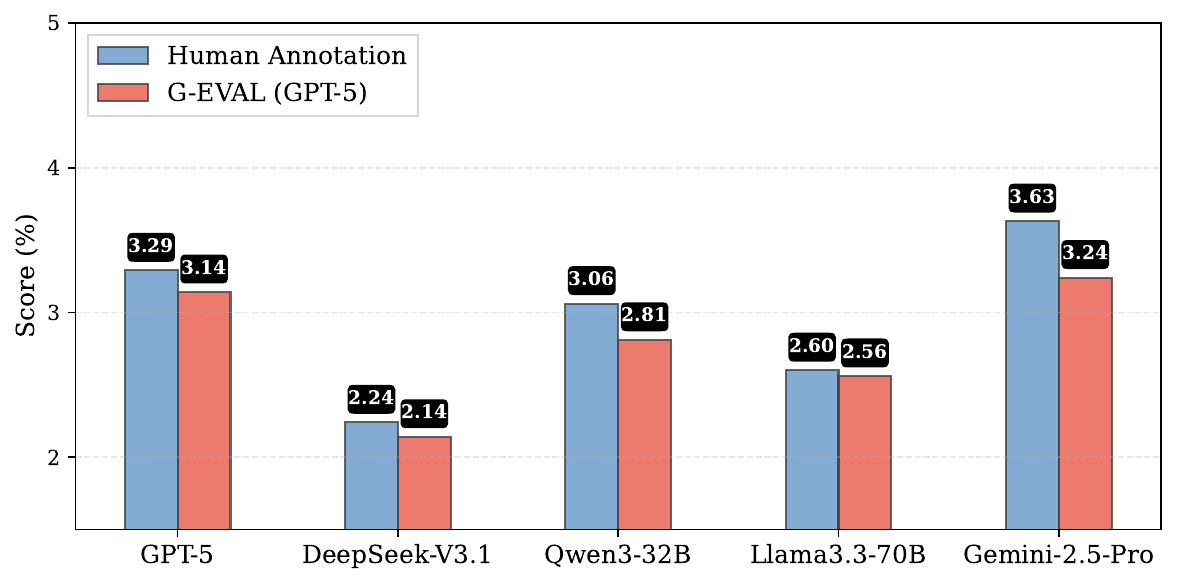}
    \caption{ Comparison of average scores of causal explanation assigned by human annotators and the G-EVAL (GPT-5) judge across five models.
    }
    \label{appendix_score_across_model}
\end{figure}

To probe potential hidden evaluator bias, we examine whether a GPT-5–based judge favors explanations generated by GPT-5 itself.
As shown in Figure~\ref{appendix_score_across_model}, we report the average scores of causal explanations assigned by human annotators and by G-EVAL based on GPT-5. These explanations are generated by five representative models. 
The results show that G-EVAL scores are consistently close to human ratings across all evaluated models. Notably, GPT-5 explanations are not favored by the GPT-5 judge, with only a 0.15-point difference between human annotations and G-EVAL, comparable to the discrepancies observed for other models. In contrast, Gemini-2.5-Pro receives higher human scores than G-EVAL, suggesting that the GPT-5 judge is more conservative when assigning high scores to explanations with complex writing styles.

\begin{figure}[!ht]
  \centering
    \includegraphics[width=1.0\linewidth]{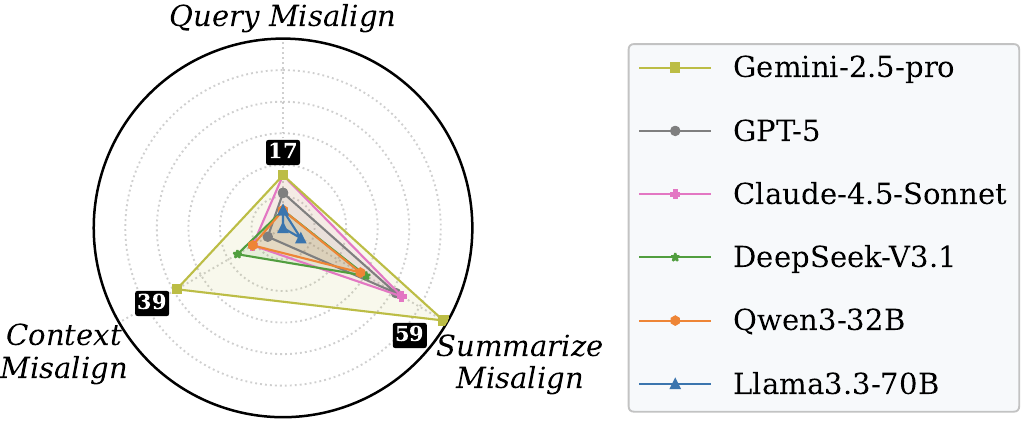}
    \caption{ Step localization accuracy of six LLM judges across three retrieval hallucination subcategories.
    }
    \label{appendix_subcategory_retrieval}
\end{figure}

\begin{figure}[!ht]
  \centering
    \includegraphics[width=1.0\linewidth]{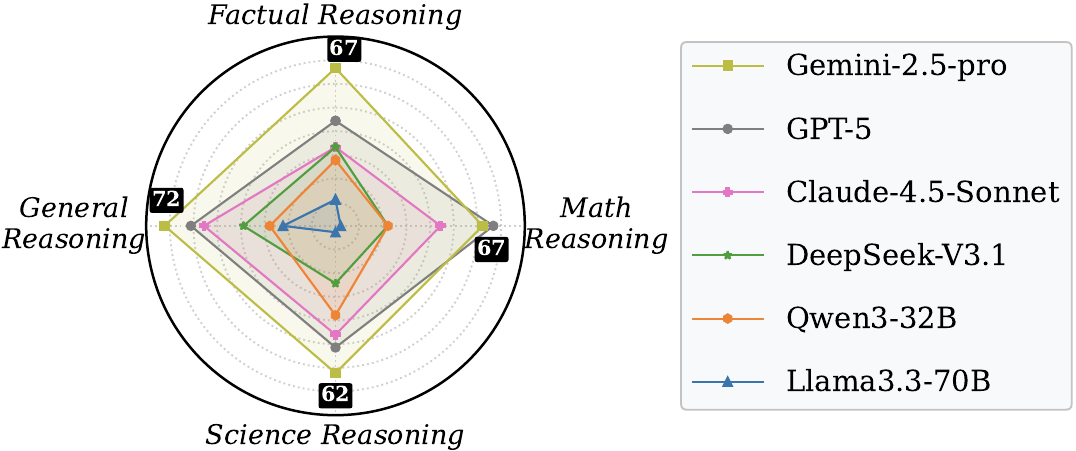}
    \caption{ Step localization accuracy of six LLM judges across four reasoning hallucination subcategories.
    }
    \label{appendix_subcategory_reason}
\end{figure}

\begin{figure}[!ht]
  \centering
    \includegraphics[width=1.0\linewidth]{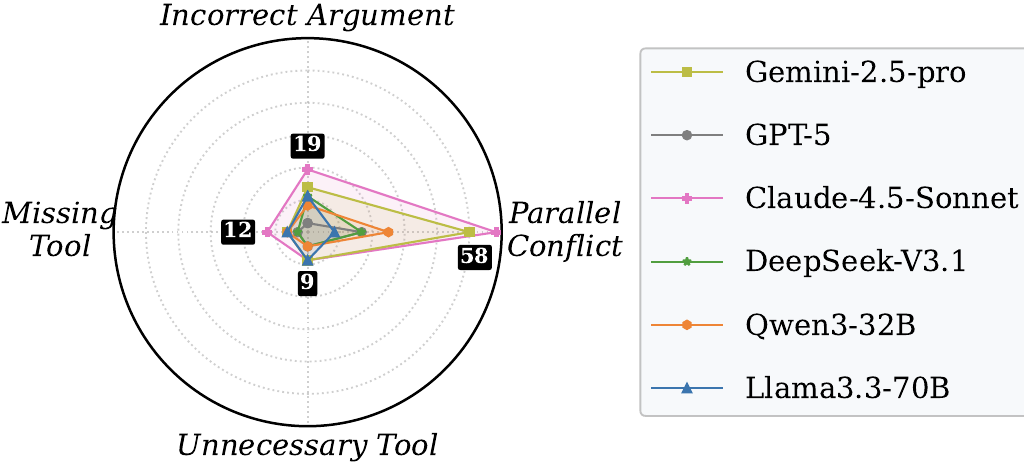}
    \caption{ Step localization accuracy of six LLM judges across four tool-use hallucination subcategories.
    }
    \label{appendix_subcategory_tool}
\end{figure}

\subsection{More Analysis on Subcategories}
\label{appendix_subcategory_analysis}
We provide subcategory-level analysis for retrieval, reasoning, and tool-use hallucinations, described as follows:
\begin{itemize}[noitemsep,left=2pt, itemsep=0pt, topsep=0pt]
\item \textit{Analysis on Retrieval Hallucination}. 
We report step localization accuracy for each retrieval hallucination subcategory in Figure~\ref{appendix_subcategory_retrieval}. The results show that Gemini-2.5-Pro achieves the strongest attribution performance across the three retrieval-hallucination subcategories. 
Notably, Gemini-2.5-Pro shows a clear advantage in summarize misalign subcategory, indicating a superior ability to localize errors introduced during evidence aggregation and compression.
In contrast, the query misalign subcategory remains the most challenging for all models, suggesting that hallucinations seeded by an incorrect retrieval intent are harder to diagnose and more likely to be confounded with later steps.

\item \textit{Analysis on Reasoning Hallucination}. We report step localization accuracy for each reasoning hallucination subcategory in Figure~\ref{appendix_subcategory_reason}. Gemini-2.5-Pro consistently attains the highest accuracy across multiple hallucination reasoning subcategories. In contrast, open-source models perform substantially worse across all subcategories, indicating limited sensitivity to logical deviations. Overall, the results suggest that accurate reasoning-hallucination attribution requires fine-grained verification of intermediate claims and their dependencies, which remains a key bottleneck for current open-source models.

\item \textit{Analysis on Tool-Use Hallucination}. We report step localization accuracy for each tool-use hallucination subcategory in Figure~\ref{appendix_subcategory_tool}. The results indicate that all models perform poorly on most subcategories, including incorrect argument, missing tool, and unnecessary tool. In contrast, Claude-4.5-Sonnet identifies parallel conflict hallucinations with notably higher accuracy, suggesting that inconsistencies from concurrent tool executions yield more explicit and verifiable contradictions.
\end{itemize}

\section{More Details on Prompt Templates}

\subsection{Templates for Question-solving Paths}
\begin{figure*}[!ht]
  \centering
    \includegraphics[width=0.95\linewidth]{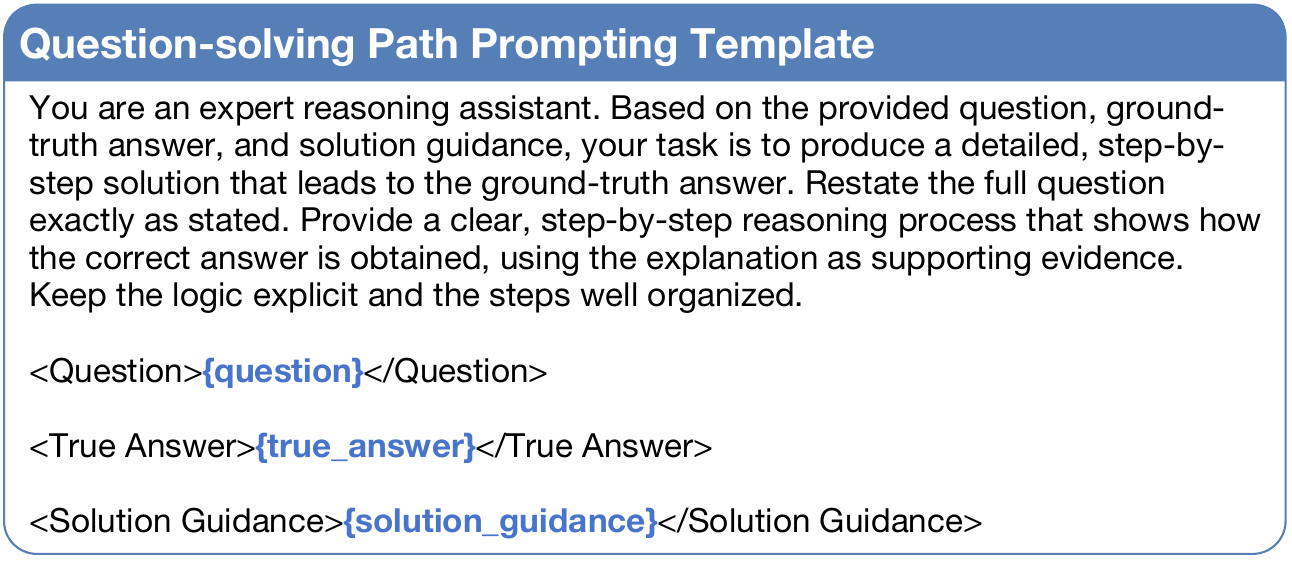}
    \caption{ Prompt template for constructing question-solving paths. The ``question'', ``true\_answer'', and ``solution\_guidance'' placeholders are replaced with the corresponding query, the ground-truth answer and the dataset-provided solution annotations.
    }
    \label{appendix_question_solving_prompt}
\end{figure*}

Figure~\ref{appendix_question_solving_prompt} illustrates the prompt template used to instruct LLMs to construct detailed question-solving paths based on the question, the ground-truth answer, and, when available, dataset-provided solution annotations.

\begin{figure*}[!ht]
  \centering
    \includegraphics[width=0.95\linewidth]{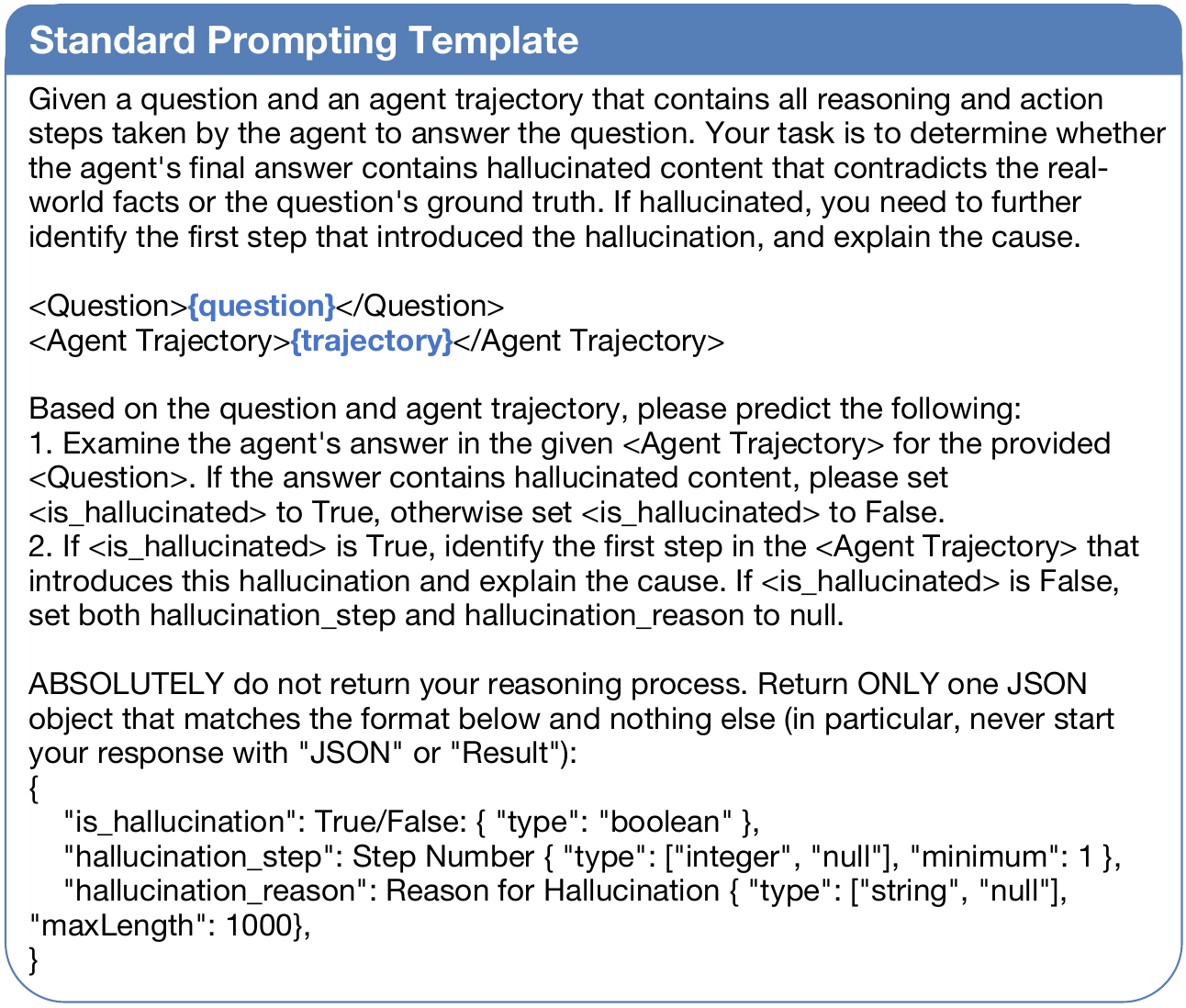}
    \caption{ Prompt template for automated hallucination judgment and attribution using standard prompting method. The ``question'' and ``trajectory'' placeholders are replaced with the corresponding query and agent trajectory to be evaluated.
    }
    \label{appendix_standard_prompting_prompt}
\end{figure*}

\begin{figure*}[!ht]
  \centering
    \includegraphics[width=1.0\linewidth]{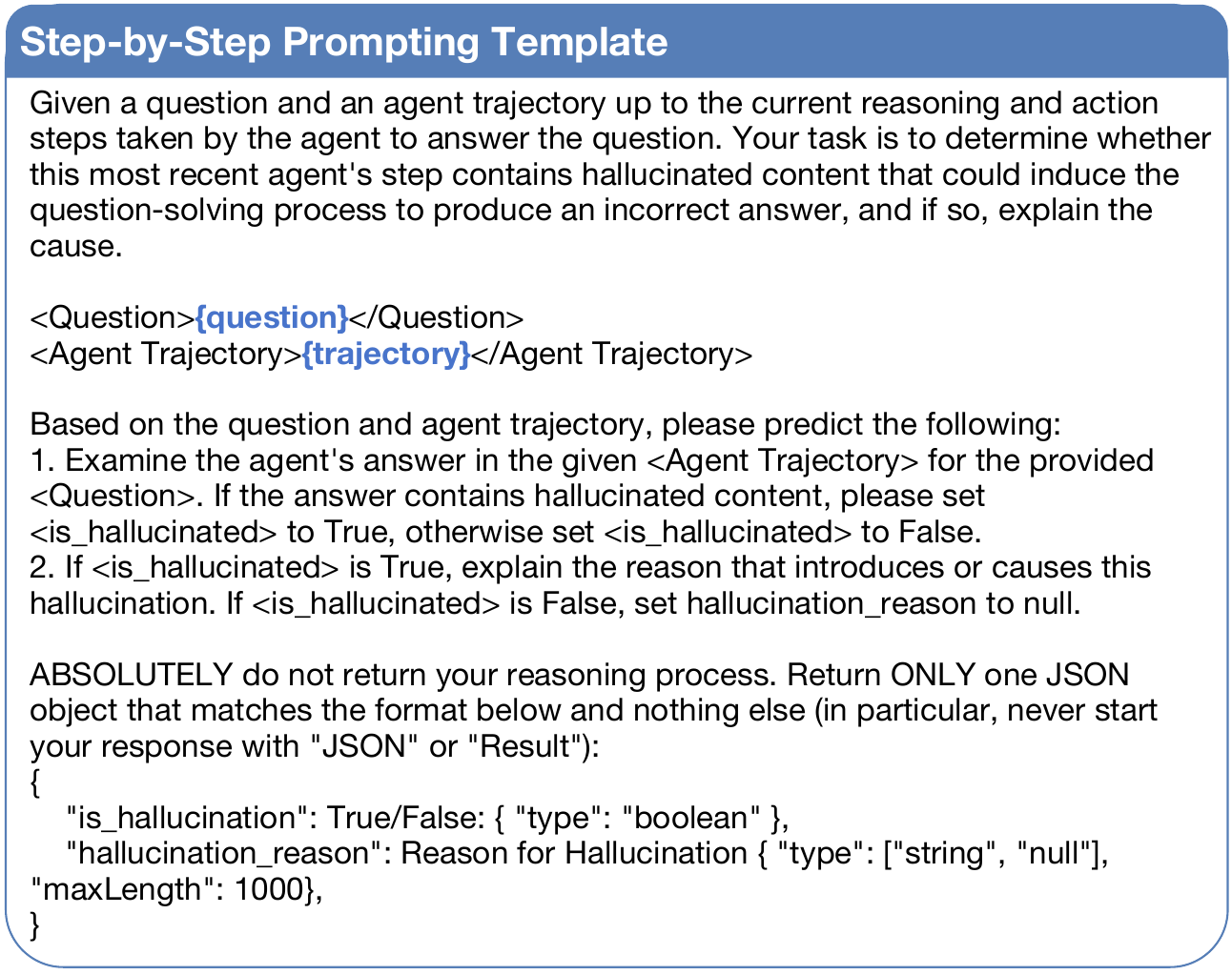}
    \caption{ Prompt template for automated hallucination judgment and attribution using step-by-step prompting method. The ``question'' and ``trajectory'' placeholders are replaced with the corresponding query and agent trajectory to be evaluated.
    }
    \label{appendix_step_by_step_prompt}
\end{figure*}

\subsection{Templates for Standard Prompting}
Figure~\ref{appendix_standard_prompting_prompt} shows the prompt template for standard prompting, where the LLM is provided with the query and the full trajectory and instructed to perform hallucination judgment and attribution.

\subsection{Templates for Step-by-Step Prompting}
Figure~\ref{appendix_step_by_step_prompt} illustrates the prompt template for step-by-step prompting, where the model processes the query and trajectory incrementally, determines whether a hallucination occurs at each step, and stops once a hallucination is detected.

\subsection{Templates for G-EVAL Evaluation}
\label{appendix_prompt_geval}
\begin{figure*}[!ht]
  \centering
    \includegraphics[width=1.0\linewidth]{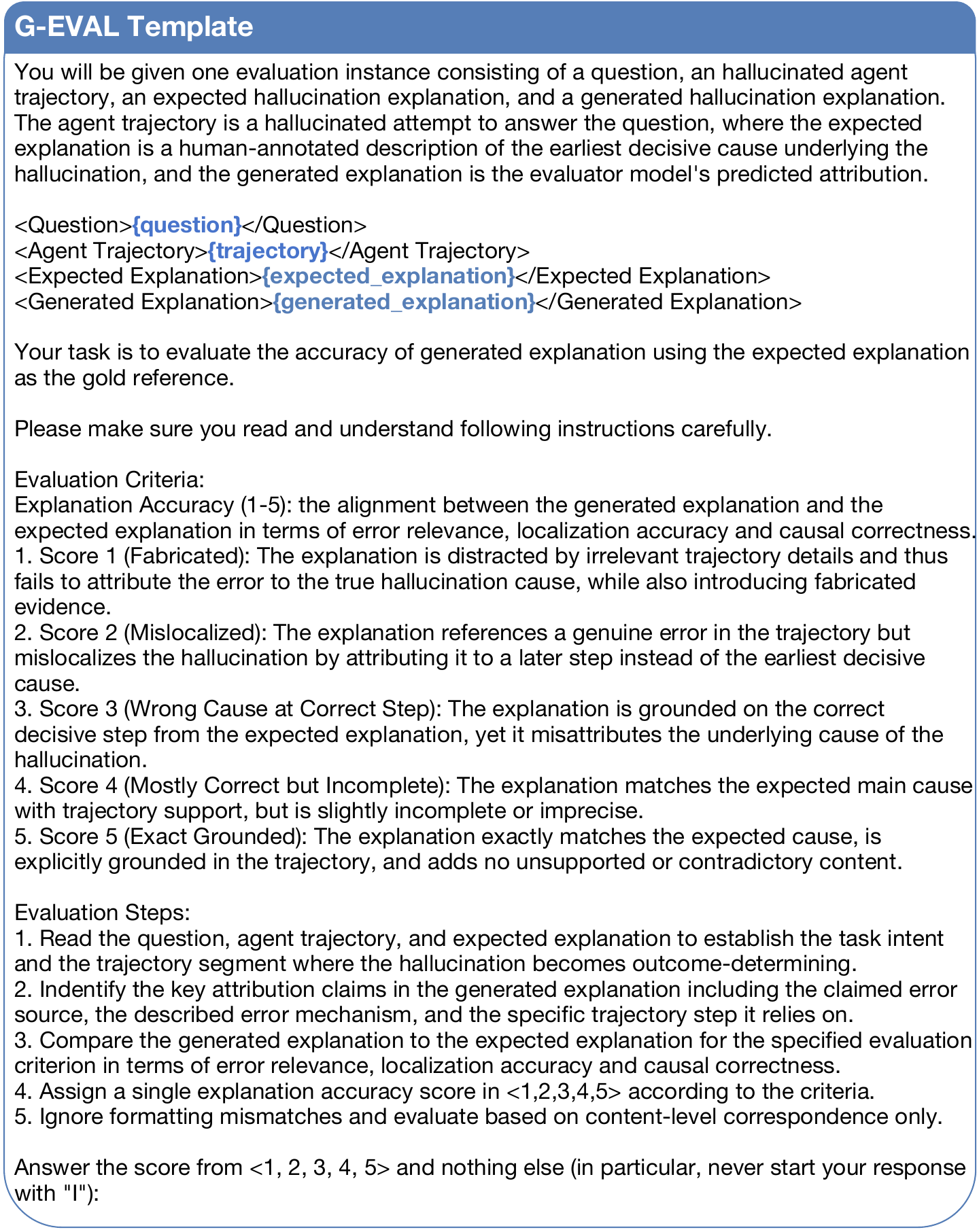}
    \caption{ Prompt template for explanation evaluation using G-EVAL method. The ``question'', ``trajectory'', ``expected\_explanation'' and ``generated\_explanation'' placeholders are replaced with the corresponding query, agent trajectory to be evaluated, the human-annotated expected explanation, and the explanation produced by the LLMs.
    }
    \label{appendix_geval_prompt}
\end{figure*}

Figure~\ref{appendix_geval_prompt} illustrates the prompt template for G-EVAL evaluation. We evaluate each model’s causal explanation, with reference to the trajectory and the human-annotated explanation. For each instance, G-EVAL assigns a five-point ordinal score under a fixed rubric that prioritizes explanation accuracy.

\section{Case Study}
\begin{figure*}[!ht]
  \centering
    \includegraphics[width=1.0\linewidth]{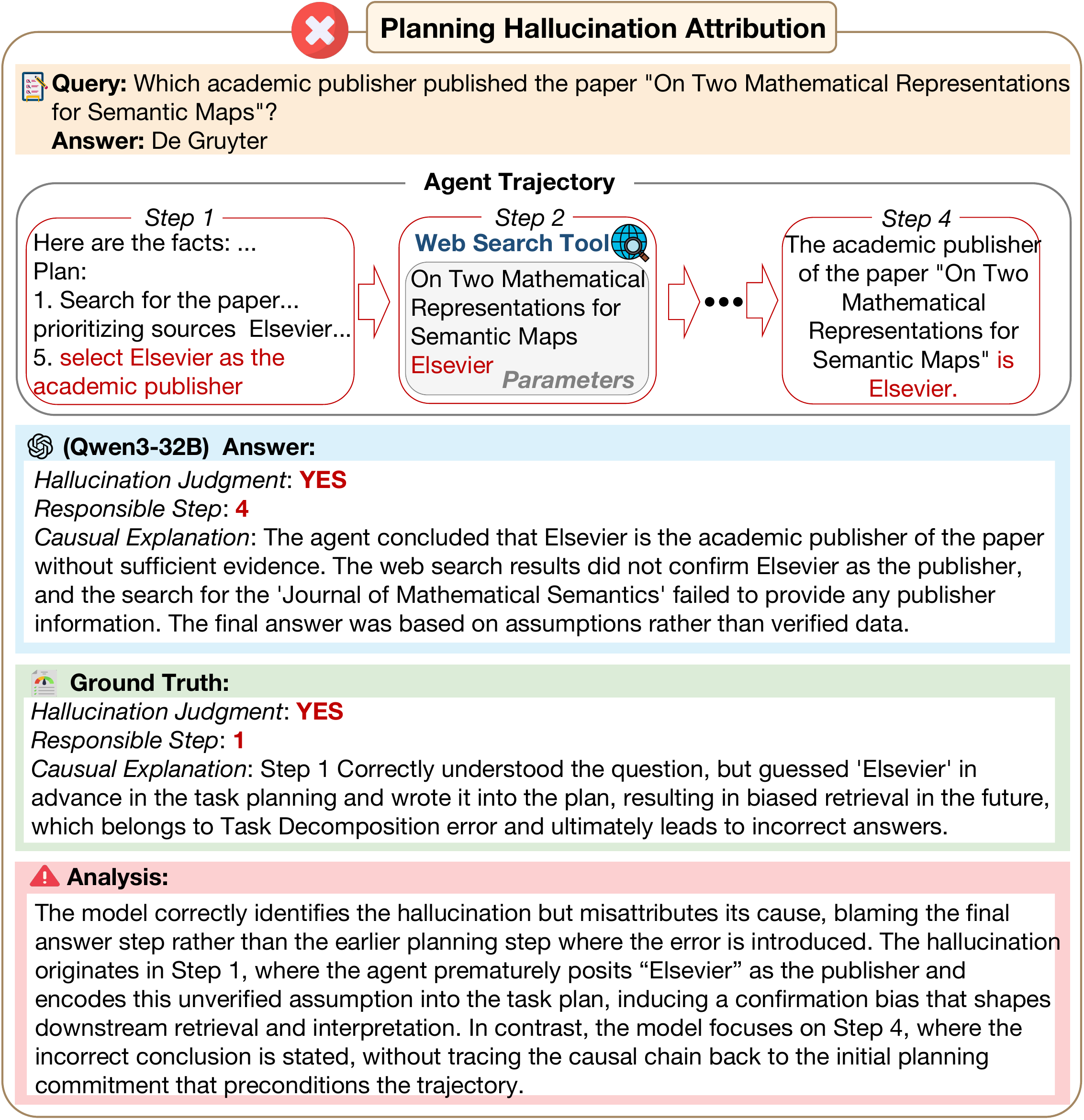}
    \caption{ Attribution example of planning hallucination category, with Qwen3-32B’s answers.
    }
    \label{appendix_case_planning}
\end{figure*}

\begin{figure*}[!ht]
  \centering
    \includegraphics[width=1.0\linewidth]{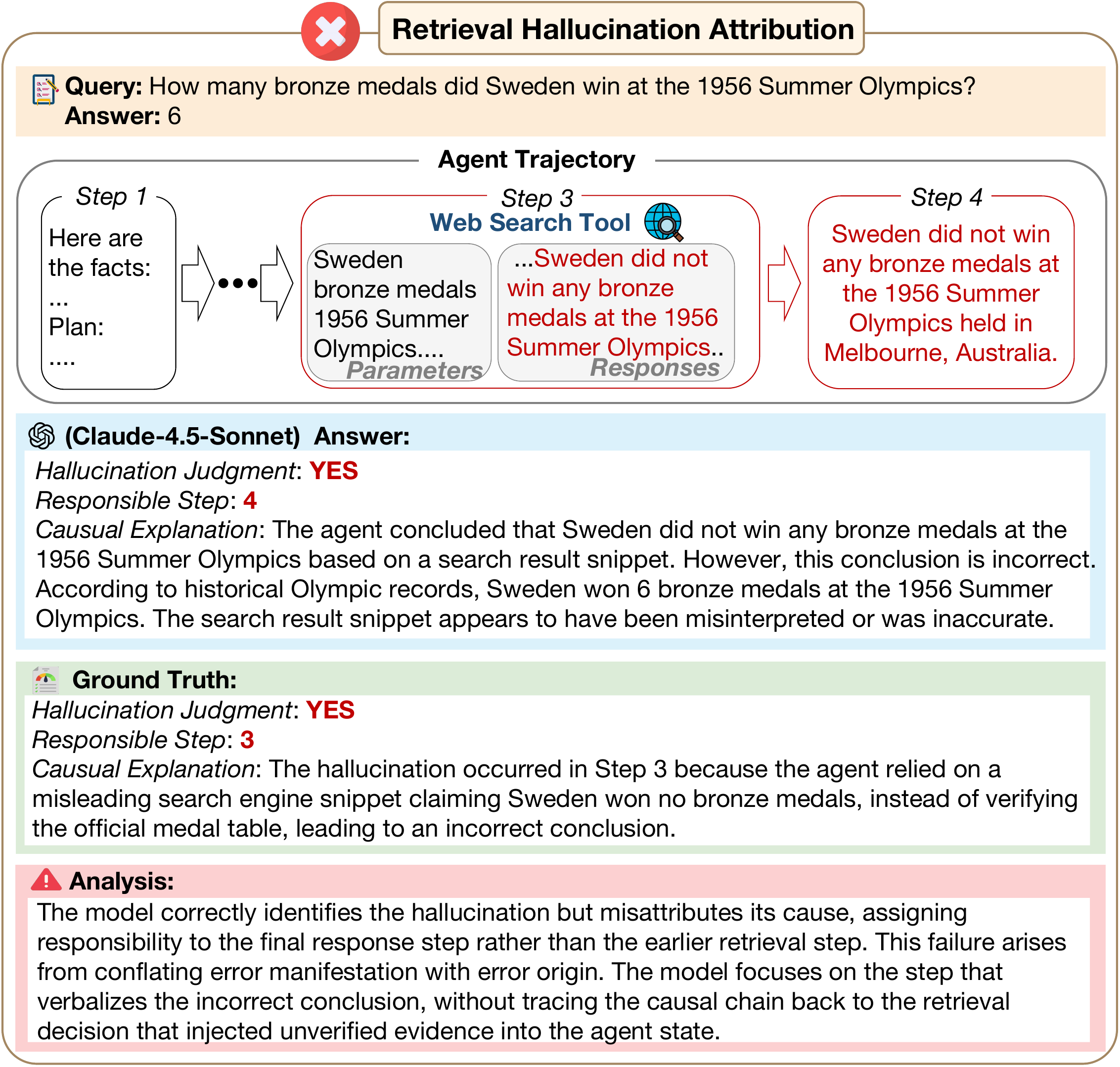}
    \caption{ Attribution example of retrieval hallucination category, with Claude-4.5-Sonnet’s answers. 
    }
    \label{appendix_case_retrieval}
\end{figure*}

\begin{figure*}[!ht]
  \centering
    \includegraphics[width=1.0\linewidth]{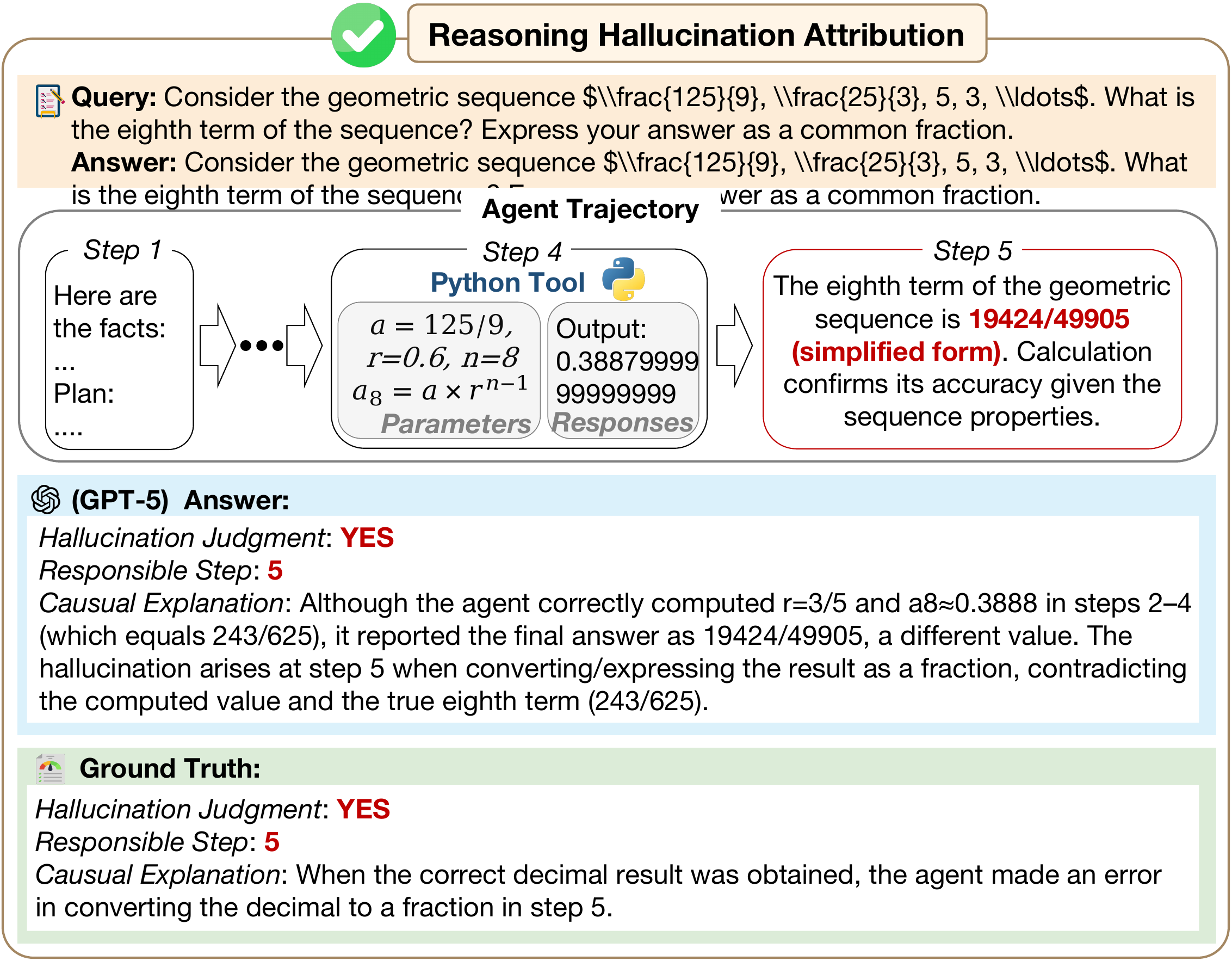}
    \caption{ Attribution example of reasoning hallucination category, with GPT-5’s answers.
    }
    \label{appendix_case_reasoning}
\end{figure*}

\begin{figure*}[!ht]
  \centering
    \includegraphics[width=1.0\linewidth]{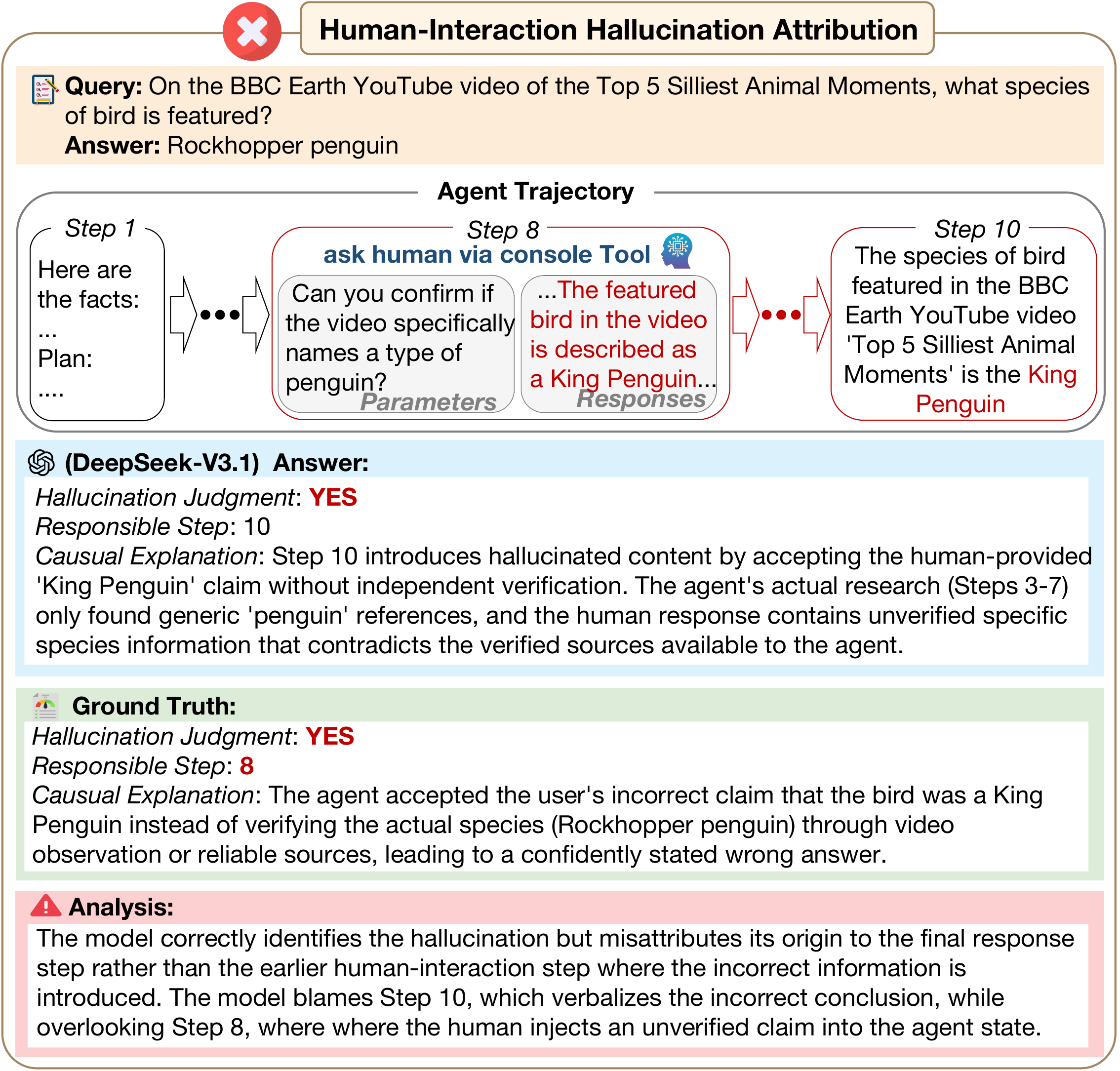}
    \caption{ Attribution example of human-interaction hallucination category, with DeepSeek-V3.1’s answers.  
    }
    \label{appendix_case_human}
\end{figure*}

\begin{figure*}[!ht]
  \centering
    \includegraphics[width=1.0\linewidth]{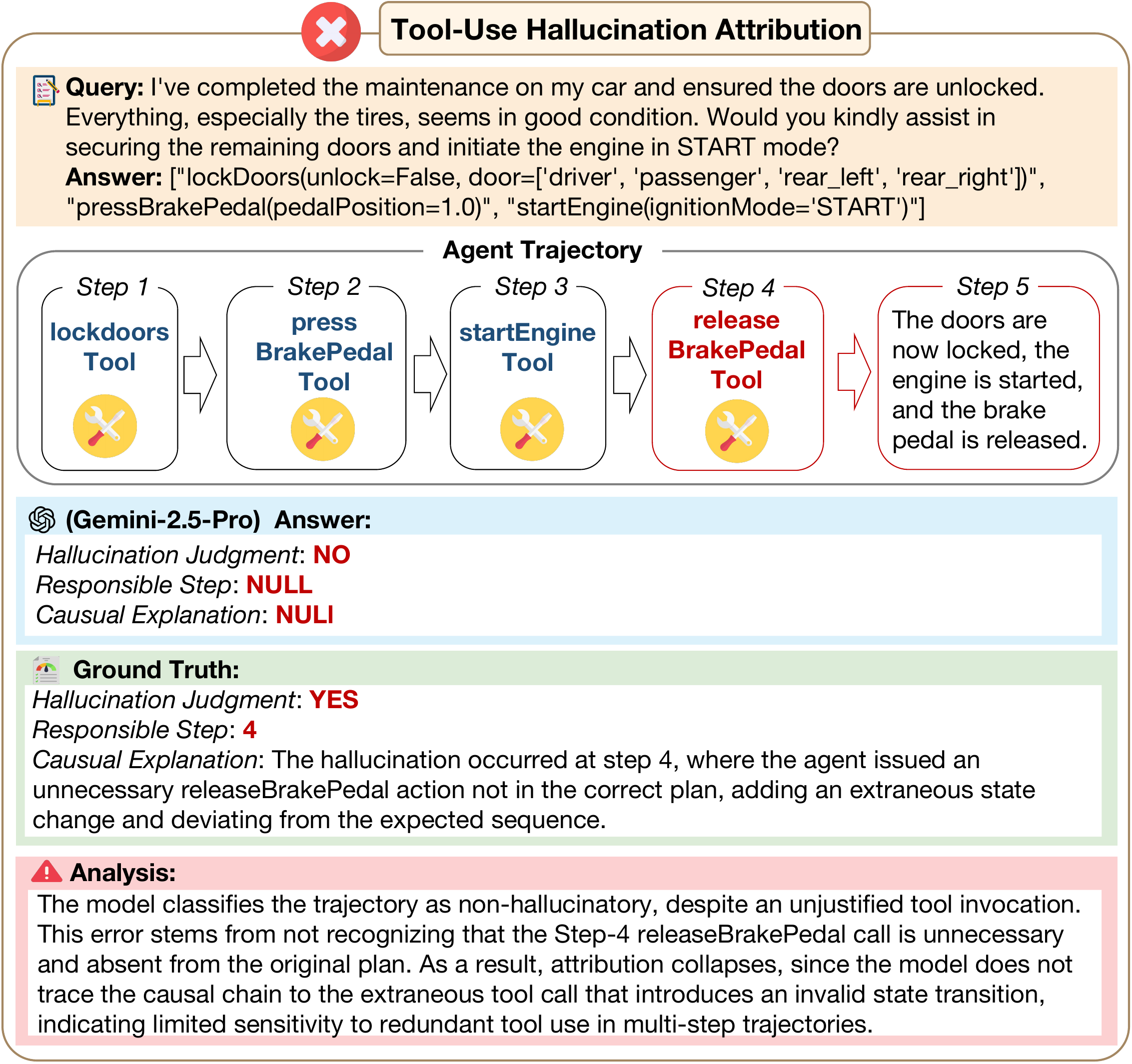}
    \caption{ Attribution example of tool-use hallucination category, with Gemini-2.5-Pro’s answers.  
    }
    \label{appendix_case_tool}
\end{figure*}

In this section, we provide qualitative case analysis of agent hallucination attribution in Figures~\ref{appendix_case_planning},~\ref{appendix_case_retrieval},~\ref{appendix_case_reasoning},~\ref{appendix_case_human},~\ref{appendix_case_tool}. This analysis is essential for assessing both hallucination identification and the ability to explain where and why hallucinations arise in agentic workflows. To this end, we examine representative hallucination cases from five models, each illustrating a dominant hallucination pattern from one category: planning, retrieval, reasoning, human interaction, or tool use. For each case, we contrast the each model’s predicted hallucination judgment, responsible step, and causal explanation with human annotations, and analyze where causal tracing breaks down.

\section{Broader Impact}
AgentHallu aims to advance the reliability and transparency of LLM-based agents by enabling systematic hallucination diagnosis in multi-step workflows. By introducing a new task of automated hallucination attribution and providing a comprehensive benchmark with fine-grained annotations, AgentHallu enables researchers to better understand where and why hallucinations arise during agent execution. This capability is critical as LLM-based agents are increasingly deployed in high-stakes applications such as healthcare, finance, and decision support, where undetected error propagation can lead to severe downstream consequences.

We acknowledge the broader societal implications of releasing benchmarks for autonomous agents. AgentHallu is constructed exclusively from publicly available data and controlled agent executions, with all annotations carefully curated to avoid sensitive or harmful content. By emphasizing transparency, reproducibility, and ethical data practices, AgentHallu fosters responsible research and deployment of LLM-based agents, contributing to the long-term realization of reliable agentic systems.

%% file: Table/category_ratio.tex
\begin{table}[h]
    \centering
    \tabcolsep=5pt
    \caption{The statistics of AgentHallu over one non-hallucination category and five hallucination categories and fourteen sub categories.}
    \label{tab:category_ratio}
    \resizebox{\linewidth}{!}{
\begin{tabular}{lrr}
            \toprule
             \textbf{Category}  &\textbf{Samples} & \textbf{Ratio(\%)} \\
             \midrule
             \rowcolor{COLOR_MEAN} 
            \hspace{0.3cm} \textbf{Non-Hallucination} &  \textbf{250} & \textbf{36.1} \\
            \midrule
            \rowcolor{category-S1} 
            \hspace{0.3cm} \textbf{Planning Hallucination} &  \textbf{67} & \textbf{9.7} \\ 
                \hspace{0.6cm} \hspace{0.3cm} $\bullet$ Fact Derive &  37 & 5.3 \\ 
                \hspace{0.6cm} \hspace{0.3cm} $\bullet$ Task Decompose &  30 & 4.3 \\ 
            \rowcolor{category-S2} 
            \hspace{0.3cm} \textbf{Retrieval Hallucination} &  \textbf{82} & \textbf{11.8} \\ 
                \hspace{0.6cm} \hspace{0.3cm} $\bullet$ Query Misalign &  18 & 2.6 \\ 
                \hspace{0.6cm} \hspace{0.3cm} $\bullet$ Context Misalign & 18 & 2.6 \\ 
                \hspace{0.6cm} \hspace{0.3cm} $\bullet$ Summarize Misalign & 46 & 4.6 \\ 
            \rowcolor{category-S3} 
            \hspace{0.3cm} \textbf{Reasoning Hallucination} &  \textbf{118} & \textbf{17.0} \\ 
                \hspace{0.6cm} \hspace{0.3cm} $\bullet$ Factual Reasoning & 18 & 2.6 \\ 
                \hspace{0.6cm} \hspace{0.3cm} $\bullet$ Science Reasoning &  37 & 5.3 \\ 
                \hspace{0.6cm} \hspace{0.3cm} $\bullet$ Math Reasoning &  45 & 6.5 \\ 
                \hspace{0.6cm} \hspace{0.3cm} $\bullet$ General Reasoning &  18 & 2.6 \\ 
             \rowcolor{category-S4} 
            \hspace{0.3cm} \textbf{Human-Interaction Hallucination} &  \textbf{73} & \textbf{10.5} \\ 
            \rowcolor{category-S5} 
            \hspace{0.3cm} \textbf{Tool-Use Hallucination} &  \textbf{103} & \textbf{14.9} \\ 
                \hspace{0.6cm} \hspace{0.3cm} $\bullet$ Missing Required Call & 32 & 4.6 \\ 
                \hspace{0.6cm} \hspace{0.3cm} $\bullet$ Incorrect Tool Arguments &  36 & 5.3 \\
                \hspace{0.6cm} \hspace{0.3cm} $\bullet$ Unnecessary Tool Call &  23 & 3.3 \\ 
                \hspace{0.6cm} \hspace{0.3cm} $\bullet$ Parallel Call Conflict &  12 & 1.6 \\ 
            \bottomrule
        \end{tabular}
    }
\end{table}

%% file: Table/agent_dataset_distribution.tex
\begin{table*}[t!]
    \centering
    \caption{Distribution of query sources across agent frameworks in AgentHallu.}
        \vspace{-0.5em}
    \label{agent_dataset_distribution}
    \resizebox{\linewidth}{!}{
    \tablestyle{3.0pt}{1.0}
    \begin{tabular}{l||cccccccc|c}
\rowcolor{COLOR_MEAN}
\toprule
\textbf{Agent Framework} & \multicolumn{1}{l}{\textbf{SimpleQA}} & \multicolumn{1}{l}{\textbf{GPQA}} & \multicolumn{1}{l}{\textbf{MATH-500}} & \multicolumn{1}{l}{\textbf{AIME2024}} & \multicolumn{1}{l}{\textbf{AIME2025}} & \multicolumn{1}{l}{\textbf{GAIA}} & \multicolumn{1}{l}{\textbf{HLE}} & \multicolumn{1}{l|}{\textbf{BFCL V3}} & \multicolumn{1}{l}{\textbf{Total}} \\ \midrule \midrule
SmolAgents               & 27                                    & 20                                & 21                                    & 1                                     & 2                                     & 14                                & 6                                & -                                     & 91                                 \\
OpenDeepSearch           & 55                                    & 9                                 & 13                                    & 3                                     & 3                                     & 10                                & 6                                & -                                     & 100                                \\
OpenManus                & 42                                    & 36                                & 8                                     & 7                                     & 7                                     & 4                                 & 0                                & -                                     & 104                                \\
Octotools                & 14                                    & 11                                & 13                                    & 0                                     & 0                                     & 5                                 & 4                                & -                                     & 47                                 \\
Magentic-One             & 28                                    & 25                                & 8                                     & 3                                     & 5                                     & 20                                & 5                                & -                                     & 94                                 \\
OWL                      & 17                                    & 36                                & 13                                    & 2                                     & 7                                     & 15                                & 3                                & -                                     & 93                                 \\ Function-Calling Agent   & -                                     & -                                 & -                                     & -                                     & -                                     & -                                 & -                                & 164                                   & 164                                \\ \midrule \midrule
Total                    & 183                                   & 137                               & 76                                    & 16                                    & 24                                    & 68                                & 25                               & -                                     & 693                                \\ \bottomrule
\end{tabular}
    }
\end{table*}

%% file: Table/agent_model_distribution.tex
\begin{table}[ht]
    \centering
    \caption{Distribution of LLM backbones across agent frameworks in AgentHallu.}
    \vspace{-0.5em}
    \label{tab:agent_model_distribution}
    \resizebox{\linewidth}{!}{
    \tablestyle{4pt}{1.03}
\begin{tabular}{lrr}
            \toprule
             \textbf{Model Backbone}  &\textbf{Samples} & \textbf{Ratio(\%)} \\
            \midrule
            \rowcolor{COLOR_MEAN} 
            \hspace{0.3cm} \textbf{SmolAgents} &  \textbf{91} & \textbf{13.1} \\ 
                \hspace{0.6cm} \hspace{0.3cm} $\bullet$ GPT-4.1 &  45 & 6.5 \\ 
                \hspace{0.6cm} \hspace{0.3cm} $\bullet$ GPT-4o &  30 & 3.2 \\ 
                \hspace{0.6cm} \hspace{0.3cm} $\bullet$ Claude-3.7-Sonnet &  5 & 0.7 \\
                \hspace{0.6cm} \hspace{0.3cm} $\bullet$ Qwen2.5-Coder-32B &  19 & 2.7 \\
            \rowcolor{COLOR_MEAN} 
            \hspace{0.3cm} \textbf{OpenDeepSearch} &  \textbf{100} & \textbf{14.4} \\ 
                \hspace{0.6cm} \hspace{0.3cm} $\bullet$ GPT-4.1 &  36 & 5.2 \\ 
                \hspace{0.6cm} \hspace{0.3cm} $\bullet$ GPT-4o & 9 & 1.3 \\ 
                \hspace{0.6cm} \hspace{0.3cm} $\bullet$ Claude-3.7-Sonnet & 19 & 2.7 \\
                \hspace{0.6cm} \hspace{0.3cm} $\bullet$ Qwen2.5-Coder-32B &  36 & 5.2 \\
            \rowcolor{COLOR_MEAN} 
            \hspace{0.3cm} \textbf{OpenManus} &  \textbf{104} & \textbf{15.0} \\ 
                \hspace{0.6cm} \hspace{0.3cm} $\bullet$ GPT-4.1 & 40 & 5.8 \\ 
                \hspace{0.6cm} \hspace{0.3cm} $\bullet$ GPT-4o &  40 & 5.8 \\ 
                \hspace{0.6cm} \hspace{0.3cm} $\bullet$ GPT-5 &  17 & 2.5 \\ 
                \hspace{0.6cm} \hspace{0.3cm} $\bullet$ Claude-3.7-Sonnet &  7 & 1.0 \\ 
             \rowcolor{COLOR_MEAN} 
            \hspace{0.3cm} \textbf{Octotools} &  \textbf{47} & \textbf{6.8} \\ 
                \hspace{0.6cm} \hspace{0.3cm} $\bullet$ GPT-4.1 &  13 & 1.9 \\ 
                \hspace{0.6cm} \hspace{0.3cm} $\bullet$ GPT-4o &  34 & 4.9 \\ 
            \rowcolor{COLOR_MEAN} 
            \hspace{0.3cm} \textbf{Magentic-One} &  \textbf{94} & \textbf{13.6} \\ 
                \hspace{0.6cm} \hspace{0.3cm} $\bullet$ GPT-4.1 & 50 & 7.2 \\ 
                \hspace{0.6cm} \hspace{0.3cm} $\bullet$ GPT-4o &  15 & 2.2 \\ 
                \hspace{0.6cm} \hspace{0.3cm} $\bullet$ GPT-5 &  21 & 3.0 \\ 
                \hspace{0.6cm} \hspace{0.3cm} $\bullet$ Claude-3.7-Sonnet &  8 & 1.2 \\ 
            \rowcolor{COLOR_MEAN} 
            \hspace{0.3cm} \textbf{OWL} &  \textbf{93} & \textbf{13.4} \\ 
                \hspace{0.6cm} \hspace{0.3cm} $\bullet$ GPT-4.1 & 72 & 10.4 \\ 
                \hspace{0.6cm} \hspace{0.3cm} $\bullet$ GPT-4o &  4 & 0.6 \\ 
                \hspace{0.6cm} \hspace{0.3cm} $\bullet$ GPT-5 &  17 & 2.5 \\ 
            \rowcolor{COLOR_MEAN} 
            \hspace{0.3cm} \textbf{Function-calling Agent} &  \textbf{164} & \textbf{23.7} \\ 
                \hspace{0.6cm} \hspace{0.3cm} $\bullet$ GPT-4.1 &  60 & 8.7 \\ 
                \hspace{0.6cm} \hspace{0.3cm} $\bullet$ Qwen3-32B &  60 & 8.7 \\ 
                \hspace{0.6cm} \hspace{0.3cm} $\bullet$ Llama3.3-70B &  44 & 6.3 \\
            \bottomrule
        \end{tabular}
    }
\end{table}

%% file: Table/model_version.tex
\begin{table*}[!t]
  \centering
  \caption{Configuration details of LLMs used for trajectory generation and hallucination evaluation in AgentHallu.}
  \label{model_version}
    \resizebox{\linewidth}{!}{
    \tablestyle{5.0pt}{1.1}
   \begin{tabular}{llccc}
\toprule
\textbf{Organization}               & \multicolumn{1}{l}{\textbf{Model}}    & \multicolumn{1}{l}{\textbf{Release}} & \textbf{Version} & \multicolumn{1}{l}{\textbf{Inference Pipeline}} \\ \midrule
\rowcolor{COLOR_MEAN}
\multicolumn{5}{l}{\textit{\textbf{Proprietary LLMs}}}                                    \\ 
\multirow{2}{*}{Google}             & \multicolumn{1}{l}{Gemini 2.5 Pro}  & 2025-6                              & \texttt{gemini-2.5-pro-06-17}                 & API                                             \\
                                    & \multicolumn{1}{l}{Gemini 2.5 Flash}    & 2025-6                                     & \texttt{gemini-2.5-flash-06-17}                   & API                                             \\
\arrayrulecolor{COLOR_MEAN}
\midrule
\arrayrulecolor{black}
\multirow{4}{*}{OpenAI}             & \multicolumn{1}{l}{GPT-5}            & 2025-8                              & \texttt{gpt-5-2025-08-07}                    & API                                             \\
                                    & \multicolumn{1}{l}{GPT-5-mini}       & 2025-8                               & \texttt{gpt-5-mini-2025-08-07}               & API                                             \\
                                    & \multicolumn{1}{l}{GPT-4.1}       & 2025-4                               & \texttt{gpt-4.1-2025-04-14}               & API                                             \\
                                    & \multicolumn{1}{l}{GPT-4o}       & 2024-12                               & \texttt{gpt-4o-2024-11-20}               & API                                             \\
\arrayrulecolor{COLOR_MEAN}
\midrule
\arrayrulecolor{black}
\multirow{2}{*}{Anthropic}          & \multicolumn{1}{l}{Claude-4.5-Sonnet} & 2025-09                              & \texttt{claude-4-5-sonnet-20250929}           & API                                                                 \\ 
                    & \multicolumn{1}{l}{Claude-3.7-Sonnet} & 2025-02                              & \texttt{claude-3-7-sonnet-20250219}           & API                                                                 \\ 
\midrule
\rowcolor{COLOR_MEAN}
\multicolumn{5}{l}{\textit{\textbf{Open-source LLMs}}}        \\
\multirow{1}{*}{DeepSeek}            & DeepSeek-V3.1                        & 2025-8                               & \texttt{deepseek-v3.1-250821}                & API                                            \\
\arrayrulecolor{COLOR_MEAN}
\midrule
\arrayrulecolor{black}         
\multirow{6}{*}{Alibaba}            & Qwen3-32B                        & 2025-4                               & \texttt{Qwen3-32B}                & Transformers                                            \\
                                    & Qwen3-8B                        & 2025-4                               & \texttt{Qwen3-8B}              & Transformers                                            \\
                                    & Qwen2.5-72B                         & 2024-9                               & \texttt{Qwen2.5-72B-Instruct}               & Transformers                                            \\
                                    & Qwen2.5-32B                          & 2024-9                               & \texttt{Qwen2.5-32B-Instruct}                & Transformers                                            \\
                                    & Qwen2.5-7B                           & 2024-9                               & \texttt{Qwen2.5-7B-Instruct}                 & Transformers                                            \\
                                    & Qwen2.5-Coder-32B                           & 2024-11                               & \texttt{Qwen2.5-Coder-32B-Instruct}                 & Transformers                                            \\
\arrayrulecolor{COLOR_MEAN}
\midrule
\arrayrulecolor{black}                                   
\multirow{2}{*}{Meta} & LLama3.3-70B                       & 2024-9                               & \texttt{Llama-3.3-70B-Instruct}                & Transformers                                    \\
                                    & LLama3.1-8B                        & 2024-7                               & \texttt{Llama-3.1-8B-Instruct}                 & Transformers                                    \\
 \bottomrule
\end{tabular}
}
\end{table*}